\documentclass{article}
\usepackage[final]{corl_2026} 
\usepackage{amsmath}
\usepackage{amssymb}
\usepackage{graphicx}
\usepackage{booktabs}
\usepackage{subcaption}
\usepackage{xcolor}
\usepackage{wrapfig2}
\usepackage{placeins}
\usepackage{enumitem}

\title{OSCAR: Obstacle Survival Curves \\ for Adaptive Robot Navigation}

\author{
Hshmat Sahak \quad Aoran Jiao \quad Nicholas Rhinehart \quad Timothy D. Barfoot\\
University of Toronto, Canada\\
{\small\texttt{\{hshmat.sahak, aoran.jiao\}@mail.utoronto.ca}}\\[-0.2ex]
{\small\texttt{\{nick.rhinehart, tim.barfoot\}@utoronto.ca}}
}

\begin{document}
\maketitle


\begin{abstract}
    A mobile robot following a graph of known routes can make costly navigation errors when a temporary obstacle blocks a critical edge: waiting too long behind a parked cart wastes time, but immediately rerouting around a person who would move in a few seconds is also inefficient. Standard reactive obstacle avoidance addresses local motion around obstacles, while fixed wait-or-reroute rules ignore how long different obstacle types tend to persist. We propose OSCAR: an adaptive survival-modeling framework for graph-based navigation with temporary blockages. Assuming obstacle class labels are available at encounter time, the robot learns class-conditioned residual clearance-time distributions from online experience, including right-censored observations when it reroutes before observing clearance. These survival models are integrated into a time-dependent graph planner that maintains obstacle memory and computes a patience threshold at each blocked edge: how long to wait before taking an alternate route. The method continuously updates its clearance estimates across episodes and uses them to balance waiting against rerouting. We evaluate the approach in simulation and on a real mobile robot in a university atrium with obstacles including people, chairs, bins, and tubes. In simulation, the learned policy's time-to-goal converges to within $1\%$ of an oracle with access to ground-truth clearance distributions after fewer than 20 observations per obstacle class, outperforming all heuristic baselines. Real-world deployment confirms that the policy improves online, adapting its patience thresholds from experience across 50 navigation episodes.
\end{abstract}



\section{Introduction}

\begin{figure*}[t]
    \centering
    \includegraphics[width=\textwidth]{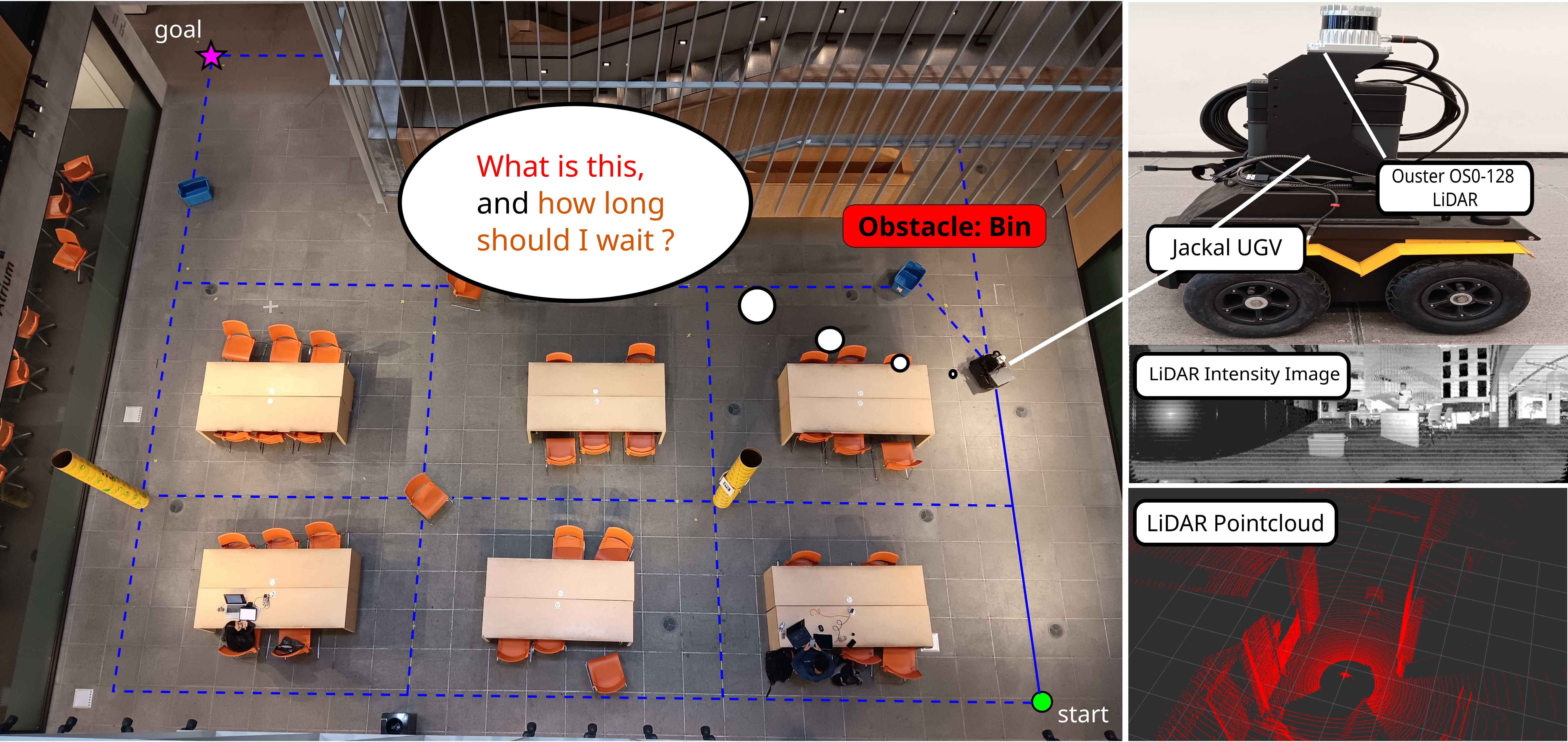}
    \captionsetup{font=footnotesize}
    \caption{
    Motivating example showing a robot navigating on a graph where it might be blocked by different objects and must decide whether to wait or reroute based on its learned survival model.
    }
    \label{fig:overview}
\end{figure*}

Robots often navigate using a graph of previously traversed or predefined routes, where vertices represent poses and edges represent traversable path segments~\citep{lacerda2019probabilistic, nardi2020long, marthi2010navigation}. This graph-based view is common when the robot should follow known routes instead of improvising around obstacles, such as indoor service robots operating in hospitals or office buildings, warehouse robots moving through designated aisles and campus delivery robots following mapped sidewalks~\citep{wurman2008coordinating,fragapane2020hospital,hawes2017strands}. In these settings, an obstacle may block an edge of the graph: a person standing in a narrow hallway, a cart left in an aisle, a chair obstructing a corridor, or a closed door along a planned route. The relevant global decision is then not simply how to avoid the obstacle locally, but whether the robot should wait for the blocked edge to become traversable again or take an alternate path through the graph. Our focus is on blockages that cannot be safely bypassed locally, as may occur in narrow corridors or designated aisles. OSCAR complements local obstacle avoidance rather than replacing it: a hierarchical navigation system can first attempt safe local avoidance and invoke OSCAR only when the route remains impassable. This decision depends on the type of obstacle and its persistence. A person may move within seconds, while a cart or chair may remain for much longer. Fixed rules such as ``always wait,'' ``always reroute,'' or obstacle-type heuristics ignore this temporal structure and can lead to unnecessary detours or excessive waiting~\citep{dieguez2008integrating, marthi2010navigation, su2005online}. Figure~\ref{fig:overview} illustrates this setting.

We formulate this problem as time-to-event prediction, where the event is obstacle clearance~\citep{kaplan1958nonparametric,cox1972regression,klein2003survival} . For each obstacle class, the robot learns a survival model describing the probability that an obstacle remains blocked as a function of time. At each encounter, the robot uses this model to compute a patience threshold: how long it should wait before rerouting. The decision compares the expected time-to-goal if the obstacle clears during the waiting period against the expected time-to-goal if the robot waits and then replans with the edge unavailable. The planner also remembers previously observed blockages, including their class and elapsed duration, making future edge costs dependent on
the planned arrival time~\citep{fujimura1995time,wellman1995path,gao2006optimal}.

As the robot operates, it updates its survival estimates from both cleared obstacles and \emph{right-censored} cases (when the robot reroutes before observing clearance)~\citep{kaplan1958nonparametric,rosen2016lifelong,tsamis2021temporal}. We assume semantic obstacle labels are available from a perception system, for example VLMs applied to LiDAR images. The resulting planner improves online while making decisions under uncertainty. 
Together, these components form \textbf{OSCAR} (\textbf{O}bstacle \textbf{S}urvival 
\textbf{C}urves for \textbf{A}daptive \textbf{R}obot navigation), a navigation policy that 
improves as the robot accumulates experience across episodes. Specifically, we contribute:
\begin{itemize}[itemsep=0pt, topsep=0pt, parsep=0pt]
    \item[(i)] \textbf{A survival-analysis formulation of obstacle persistence} for 
    graph-based robot navigation, to our knowledge the first, which enables 
    principled handling of right-censored observations that arise naturally when 
    the robot reroutes before an obstacle clears;
    
    \item[(ii)] \textbf{A local wait-or-reroute decision rule} that selects a patience 
    threshold at a blocked edge by minimizing expected time-to-goal under the current 
    survival estimate, validated in simulation where the learned policy's mean time-to-goal across 100 random seeds converges to within 1\% of an oracle with access to ground-truth clearance distributions; and
    
    \item[(iii)] \textbf{A time-dependent global planner with obstacle memory} that 
    incorporates survival estimates as arrival-time-dependent edge costs across the 
    full route graph, with FIFO edge-arrival functions that preserve the applicability 
    of Dijkstra's algorithm despite history-dependent edge costs, validated on a robot navigating a campus atrium across 50 episodes.
\end{itemize}

\section{Related Work}
\noindent\textbf{Dynamic Obstacle Avoidance and Social Navigation.}
A large body of work on robot navigation in dynamic environments focuses on local obstacle avoidance, where decisions are made reactively over short time horizons using current sensor observations~\citep{fox1997dynamic,kruse2013human,trautman2015robot,ratatouille}. 
These methods are effective for short-horizon motion control, but they typically operate at the trajectory or control level rather than at the level of global route choice. In contrast, our work addresses a complementary decision problem: when a robot encounters a blockage along a graph-based route, how long should it wait for the obstacle to clear before abandoning that edge to take a different path?

\noindent\textbf{Planning under Uncertainty and the Canadian Traveller Problem.}
Global navigation is commonly formulated as graph search, with methods such as Dijkstra's algorithm and A* computing shortest paths over vertices and traversable edges~\citep{dijkstra1959note,hart1968formal}. Extensions to uncertain settings include stochastic shortest-path formulations, Markov decision processes, and variants of the Canadian Traveller Problem (CTP), where edge costs or traversability may be unknown~\citep{papadimitriou1989shortest,nikolova2008route,eyerich2010high,guo2019robust}. CTP is especially close to our setting because edge blockages are revealed online as the robot traverses the graph, rather than being known before planning. However, standard CTP-style formulations typically treat blockages as static hidden variables or one-time realizations of traversability. In contrast, our setting requires reasoning about persistence: once a robot observes a blockage, it must reason about the likelihood that it will clear within a useful time horizon.

\noindent\textbf{Time-Dependent Routing.}
Time-dependent shortest-path methods allow edge costs to depend on the time at which an edge is reached~\citep{halpern1977shortest, hall1986fastest,orda1991minimum,wellman1995path}. When the resulting edge arrival functions satisfy the FIFO property, meaning that departing later cannot lead to arriving earlier on the same edge, variants of Dijkstra's algorithm can still be applied~\citep{halpern1977shortest,orda1991minimum,fujimura1995time}. This structure is relevant to our setting because obstacle memory makes edge costs time-dependent: the expected future wait on edges we previously observed blocked and decided to reroute depends on the robot's planned arrival time~\citep{marthi2010navigation,wellman1995path,gao2006optimal}. Arriving later may reduce the expected remaining blockage time, but the induced arrival-time function remains FIFO. We therefore incorporate learned obstacle-persistence estimates into graph search by inflating edge costs with planned-arrival-dependent expected waits.

\noindent\textbf{Survival Modeling for Obstacle Persistence.}
Survival analysis models the time until an event occurs while naturally handling censored observations~\citep{kaplan1958nonparametric,cox1972regression,klein2003survival}. In our setting, the event is an obstacle clearing, and we fit a non-parametric Kaplan--Meier (KM) survival model for each obstacle class. These class-conditioned clearance distributions support planning in two ways: for the currently encountered obstacle, they estimate the probability of clearance within a candidate patience threshold; for previously observed obstacles elsewhere in the graph, they inform the expected wait at the robot's planned arrival time. 

\section{OSCAR: Survival-Guided Wait-or-Reroute Planning}
\label{sec:method}
We consider a robot navigating on a graph \(G=(V,E)\), where vertices represent robot poses and edges represent feasible motions with deterministic traversal times \(w(e)\ge 0\). Given start and goal vertices \(s,g\in V\), each \emph{episode} consists of one traversal from \(s\) to \(g\). Edges may be temporarily blocked by dynamic obstacles that appear randomly. Class labels are provided at encounter time.

\begin{figure}[htbp]
\vspace{-0.8em}
\centering
    \includegraphics[width=\linewidth]{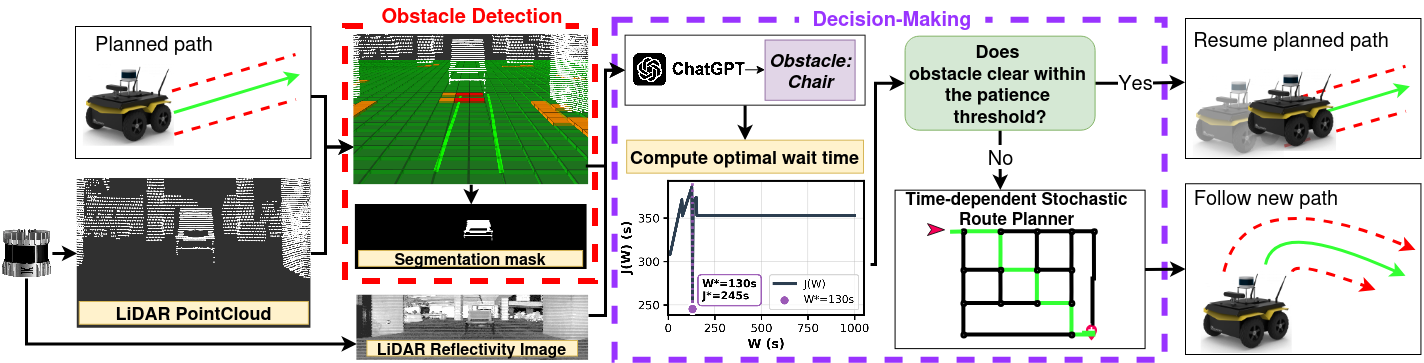}
    \captionsetup{font=footnotesize}
    \caption{
Our proposed pipeline. The robot follows the planned path till a blockage is detected. The reflectivity image and segmentation mask are used to determine the object type. The planner then uses the corresponding survival model to compute a patience threshold \(W^\star\), waits up to that threshold, and either resumes the original path if the obstacle clears or replans if the blockage persists.
    }
    \label{fig:pipeline}
\end{figure}

Our objective is to learn a navigation policy that minimizes expected time-to-goal over repeated episodes. At each blocked-edge encounter, the policy selects a patience threshold \(W^\star\) using its current estimates of obstacle statistics. The robot waits up to \(W^\star\), and if the obstacle remains blocked after that, takes another path. Since we do not assume access to an obstacle association module, repeated observations of a blocked edge at different times are treated as separate encounters. With more experience, the learned policy should approach an oracle with access to the ground-truth clearance distributions, blockage probability \(p_{\mathrm{block}}\), and object type probabilities given blocked, \(p_k\).

Figure~\ref{fig:pipeline} shows the full pipeline. In this section, we focus on the decision-making module. The module needs three quantities: a survival model that predicts how long each obstacle class tends to remain, an edge cost that converts these predictions into expected navigation delay, and a local wait-or-reroute rule that chooses the patience threshold \(W^\star\) when the robot reaches a blocked edge.

\subsection{Learning Class-Conditioned Obstacle Persistence}

For each obstacle class \(k\in\mathcal{K}\), we model how long a blockage remains after the robot observes it. When the robot reaches an edge and detects a class-\(k\) blockage, let \(R^{(k)}\) be the remaining time until that blockage clears. The class-conditioned survival function is
$
S_k(\tau)=\mathbb{P}(R^{(k)}>\tau),
$
which gives the probability that a newly observed class-\(k\) blockage is still present after waiting \(\tau\) seconds.

We estimate one survival curve per obstacle class using the Kaplan--Meier (KM)
estimator~\citep{kaplan1958nonparametric,klein2003survival}, which supports \emph{learning from both clearance and ``right-censored'' observations}. A right-censored observation occurs when
the robot reroutes before observing the obstacle clear, so \emph{it only observes that the
clearance time exceeded the observed waiting time, rather than the true clearance time}. For class \(k\), let \(0<\tau_1<\tau_2<\cdots<\tau_J\) be the distinct observed clearance times. If \(d_j\) is the number of clearances at time \(\tau_j\) and \(n_j\) is the number of samples still under observation immediately before \(\tau_j\), then:
\begin{equation}
\widehat{S}_k(\tau)
=
\prod_{\tau_j\le \tau}\left(1-\frac{d_j}{n_j}\right).
\end{equation}

\begin{wrapfigure}{0}{0.66\linewidth}
    \vspace{-0.8em}
    \centering
    \includegraphics[width=\linewidth]{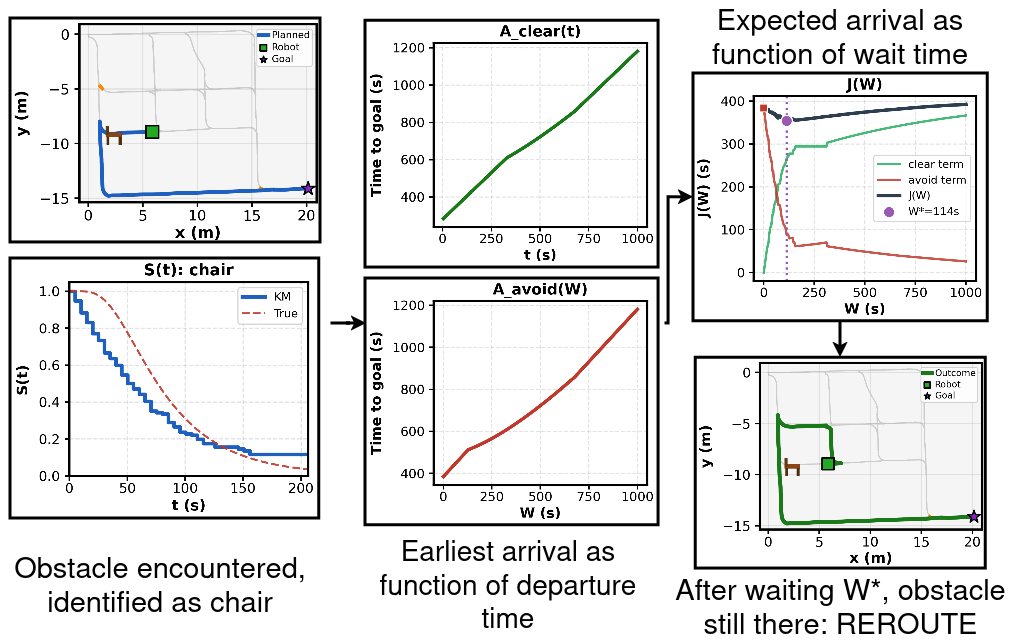}
    \vspace{-0.8em}
    \captionsetup{font=footnotesize}
    \caption{
    Example wait-or-reroute decision at a blocked edge. After identifying a chair, the planner uses the learned survival model to compute \(A_{\mathrm{clear}}(c)\), \(A_{\mathrm{avoid}}(W)\), and \(J(W)\). The minimizer \(W^\star\) gives the patience threshold; here the obstacle persists, so the robot reroutes.
    }
    \label{fig:sample_decision}
\end{wrapfigure}

At each observed clearance time $\tau_j$, the factor $1-d_j/n_j$ estimates the fraction of obstacles still blocked among those at risk. Right-censored observations contribute to the risk set until their last observation, after which they are removed without being counted as clearance events. The fitted survival curve is piecewise constant, with probability mass $\widehat{p}_{k,j} = \widehat{S}_k(\tau_{j-1})-\widehat{S}_k(\tau_j)$ assigned to event time \(\tau_j\). Figure~\ref{fig:sample_decision} includes an example KM fit. 

\subsection{Arrival-Time-Dependent Edge Costs}

During planning, the robot must estimate how costly each edge will be if reached at a future time \(t\). We write this cost as
$
\widehat{c}(e,t)=w(e)+\widehat{\Delta}(e,t),
$
where \(w(e)\) is the nominal traversal time and \(\widehat{\Delta}(e,t)\) is the expected extra delay caused by obstacles. We first consider an edge with no remembered blockage. In this case, the robot only accounts for the possibility that a new obstacle will be discovered when it arrives. Let
$
\widehat{p}_{\mathrm{block}}
=
\frac{\#\text{ blocked-edge encounters}}{\#\text{ edge traversal attempts}}$, $
\widehat{p}_k
=
\frac{\#\text{ observations of class }k}{\#\text{ total obstacle observations}},
$
and let
$
\widehat{\mu}_k
=
\int_0^\infty \widehat{S}_k(u)\,du
$
be the estimated mean remaining clearance time for a newly observed class-\(k\) obstacle. This follows from the survival identity
\(\mathbb{E}[R]=\int_0^\infty \mathbb{P}(R>u)\,du\). Therefore, the expected delay from a possible new blockage is
$
\widehat{\Delta}_{\mathrm{new}}
=
\widehat{p}_{\mathrm{block}}
\sum_{k\in\mathcal{K}} \widehat{p}_k\,\widehat{\mu}_k .
$
Thus, if edge \(e\) has no remembered obstacle state, we use
$
\widehat{\Delta}(e,t)=\widehat{\Delta}_{\mathrm{new}}.
$

\textbf{Edges with remembered obstacle state:}
If edge \(e\) was previously observed to be blocked, the robot has more information than the prior blockage rate. It stores
\(M(e)=(k_e,t_{\mathrm{first}}(e),t_{\mathrm{last}}(e))\), where \(k_e\) is the observed class, \(t_{\mathrm{first}}\) is the first observation time, and \(t_{\mathrm{last}}\) is the most recent time the obstacle was confirmed to still be present. Suppose the planner expects to reach this edge at time \(t\ge t_{\mathrm{last}}(e)\). The obstacle was already known to survive until age
$
a_e=t_{\mathrm{last}}(e)-t_{\mathrm{first}}(e),
$
and would have age
$
b_e(t)=t-t_{\mathrm{first}}(e)
$
at the planned arrival time. The probability that the same obstacle is still present when the robot arrives is therefore
$
q_e(t)
=
\mathbb{P}(R^{(k_e)}>b_e(t)\mid R^{(k_e)}>a_e)
=
\frac{\widehat{S}_{k_e}(b_e(t))}{\widehat{S}_{k_e}(a_e)}.
$
If it is still present at arrival, the expected additional wait is the mean residual life after already lasting \(b_e(t)\) seconds:
$
\widehat{\mu}^{\mathrm{old}}_e(t)
=
\int_0^\infty
\frac{\widehat{S}_{k_e}(b_e(t)+u)}
     {\widehat{S}_{k_e}(b_e(t))}\,du .
$
Thus the edge delay is a mixture of two cases: with probability \(q_e(t)\), the old obstacle is still present and causes residual delay; otherwise, the old obstacle has cleared and the edge is treated like an ordinary edge that may contain a new blockage:
$
\widehat{\Delta}(e,t)
=
q_e(t)\,\widehat{\mu}^{\mathrm{old}}_e(t)
+
\bigl(1-q_e(t)\bigr)\widehat{\Delta}_{\mathrm{new}}.
$

\subsection{Wait-or-Reroute Decision at a Blocked Edge}

When the robot actually reaches a blocked edge, it must choose how long to wait before abandoning that edge. A patience threshold \(W\) means: wait up to \(W\) seconds; if the obstacle clears before then, replan with the edge available; otherwise, replan with the blocked edge forbidden. Define \(A_{\mathrm{clear}}(c)\) as the remaining time-to-goal if the obstacle clears after waiting \(c\) seconds. Define \(A_{\mathrm{avoid}}(W)\) as the remaining time-to-goal if the robot waits until \(W\), the obstacle is still present, and the robot reroutes with that edge forbidden. In both cases, remaining time is measured from encounter time.

The expected remaining time-to-goal under threshold \(W\) is:
\begin{equation}
J(W)
=
\int_0^W A_{\mathrm{clear}}(c)\,dF_k(c)
+
S_k(W)\,A_{\mathrm{avoid}}(W),
\end{equation}

where \(F_k(c)=1-S_k(c)\). The first term averages over all ways the obstacle might clear before the deadline \(W\). The second term handles the event that the obstacle survives past \(W\), in which case the robot gives up and reroutes. We choose
$
W^\star = \arg\min_{W\in[0,W_{\max}]} J(W).    
$
This decision-making process is illustrated in Figure~\ref{fig:sample_decision}. In the learned method, we replace \(S_k\) with the KM estimate \(\widehat{S}_k\). Since the fitted curve is piecewise constant, the integral becomes a finite sum over observed clearance times:
\begin{equation}
\widehat{J}(W)
=
\sum_{\tau_j \le W}
\widehat{p}_{k,j}\,A_{\mathrm{clear}}(\tau_j)
+
\widehat{S}_k(W)\,A_{\mathrm{avoid}}(W).
\end{equation}

\subsubsection{Computing \(A_{\mathrm{clear}}\) and \(A_{\mathrm{avoid}}\)}

Both quantities are shortest-path costs under different assumptions about the blocked edge. For \(A_{\mathrm{clear}}(c)\), the edge becomes available after \(c\) seconds, although the planner is not forced to use it. For \(A_{\mathrm{avoid}}(W)\), the robot waits until \(W\) and then plans with the blocked edge removed. Because other edges may also have arrival-time-dependent delays, these are time-dependent shortest-path problems. In ~\ref{app:functional_planner}, we describe a functional Bellman--Ford procedure that propagates earliest-arrival functions over the graph. In the main algorithm, we use the fact that KM survival curves are piecewise constant and the edge costs satisfy FIFO: probability terms only change at KM event times, and waiting cannot improve arrival time except through those event changes. Thus, it suffices to evaluate \(\widehat{J}(W)\) on the finite candidate set of observed event times, \(0\), and \(W_{\max}\). For each candidate, we compute \(A_{\mathrm{clear}}\) and \(A_{\mathrm{avoid}}\) using Dijkstra with the time-dependent edge costs \(\widehat{c}(e,t)\).

\subsection{Online Adaptation}

After each blocked-edge encounter, the robot records
\(\{t=\min(R^{(k)},W^\star), \delta=\mathbf{1}[R^{(k)}\le W^\star]\}\), where \(t\) is the amount of time actually observed and \(\delta\) indicates whether the obstacle cleared before the robot rerouted. This sample updates the KM curve \(\widehat{S}_k\) for the observed class. The robot also updates \(\widehat{p}_{\mathrm{block}}\) from edge traversal attempts and \(\widehat{p}_k\) from observed obstacle classes.

\section{Experimental Evaluation of Survival-Guided Navigation}
\label{sec:result}
\vspace{-0.8em}

In simulation, we verify our approach under controlled obstacle-generation and per-class clearance distributions. Real-world experiments serve as hardware validation on an indoor navigation task.

\subsection{Simulation Experiments}
We evaluate the method in simulation under controlled obstacle dynamics. Obstacles are generated independently of the robot according to a Poisson process: each spawn time is sampled from the process, an edge is selected uniformly at random, an obstacle class \(k \in \{\text{person}, \text{chair}, \text{bin}, \text{tube}\}\) is sampled with probabilities 55\%, 30\%, 10\% and 5\%, respectively, and a lifetime is sampled from the corresponding class-conditioned clearance-time distribution that we define. The Poisson rate is chosen to obtain a desired steady-state edge blockage probability, \(5\%\). Full details in ~\ref{app:simulation_details}. 

\begin{figure}[htbp]
    \centering
    \includegraphics[width=\linewidth]{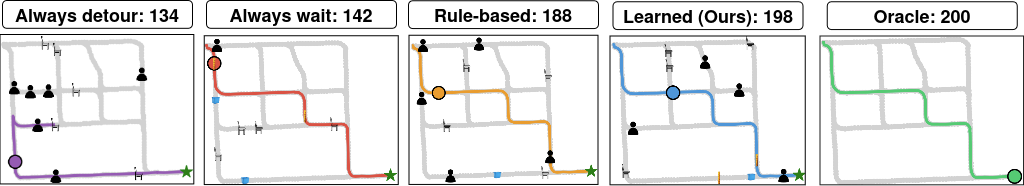}
    \captionsetup{font=footnotesize}
    \caption{Race to 200. Oracle comes first place. Our method is best amongst alternatives (198).}
    \label{fig:horse_race}
\end{figure}

We compare the proposed learned KM policy against four baselines: \emph{Always Wait}, \emph{Always Reroute}, \emph{Rule-Based} (wait for people, reroute for rest), and \emph{Greedy CTP}. The latter is a greedy CTP-style baseline that immediately replans when a blockage is observed and permanently forbids the blocked edge for the remainder of the episode. We also include an \emph{Oracle} version of our method that uses the true clearance distributions, \(p_{\mathrm{block}}\), and \(p_k\). The oracle is evaluated using the exact time-dependent formulation with Bellman--Ford, while the learned policy uses the KM discretization and evaluates the finite candidate set by running Dijkstra for both \(A_{\mathrm{clear}}\) and \(A_{\mathrm{avoid}}\) at each candidate time. An episode is counted as successful if the robot reaches the goal within \(1\,\mathrm{hr}\). The simulation visualization, as well as scores in a race till one reaches 200 successful episodes are shown in Figure~\ref{fig:horse_race}.

\begin{figure}[htbp]
\vspace{-0.8em}
    \centering
    \begin{subfigure}[t]{0.49\linewidth}
        \centering
        \includegraphics[
    width=\linewidth,
    height=0.65\linewidth,
    trim={0 7pt 0 0},
    clip
]{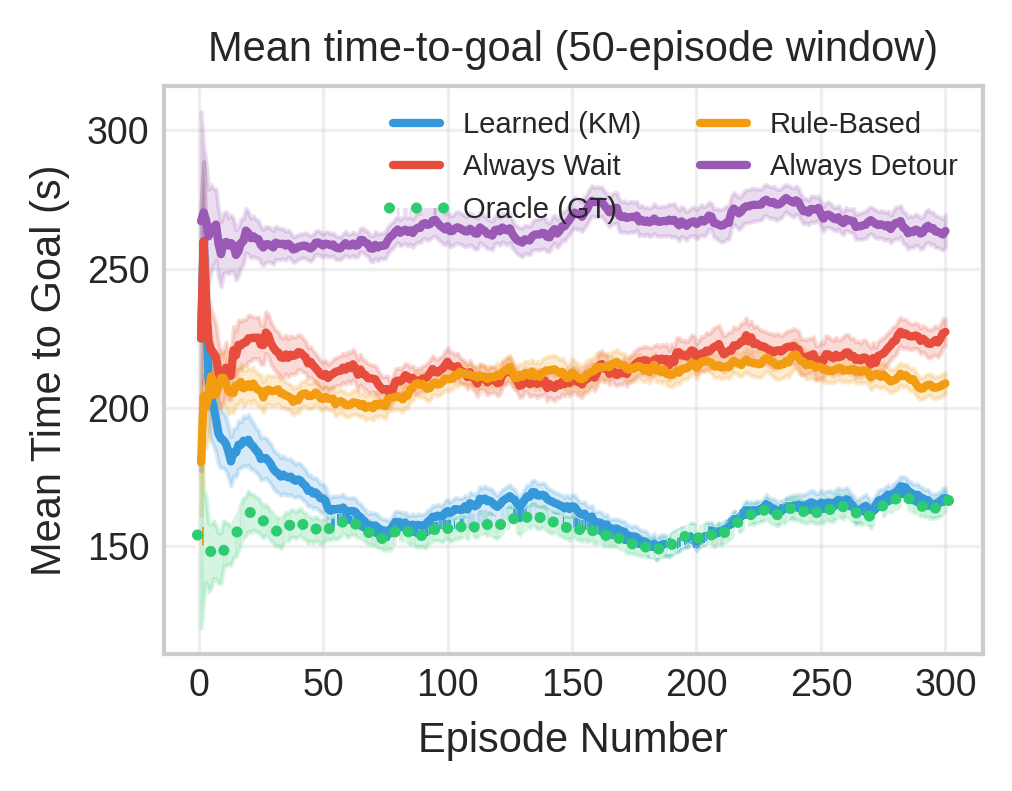}
        \caption{Mean time-to-goal across episodes.}
        \label{fig:sim_results}
    \end{subfigure}
    \hfill
    \begin{subfigure}[t]{0.49\linewidth}
        \centering
        \includegraphics[width=\linewidth,
        height=0.65\linewidth,
    trim={0 7pt 0 0},
    clip]{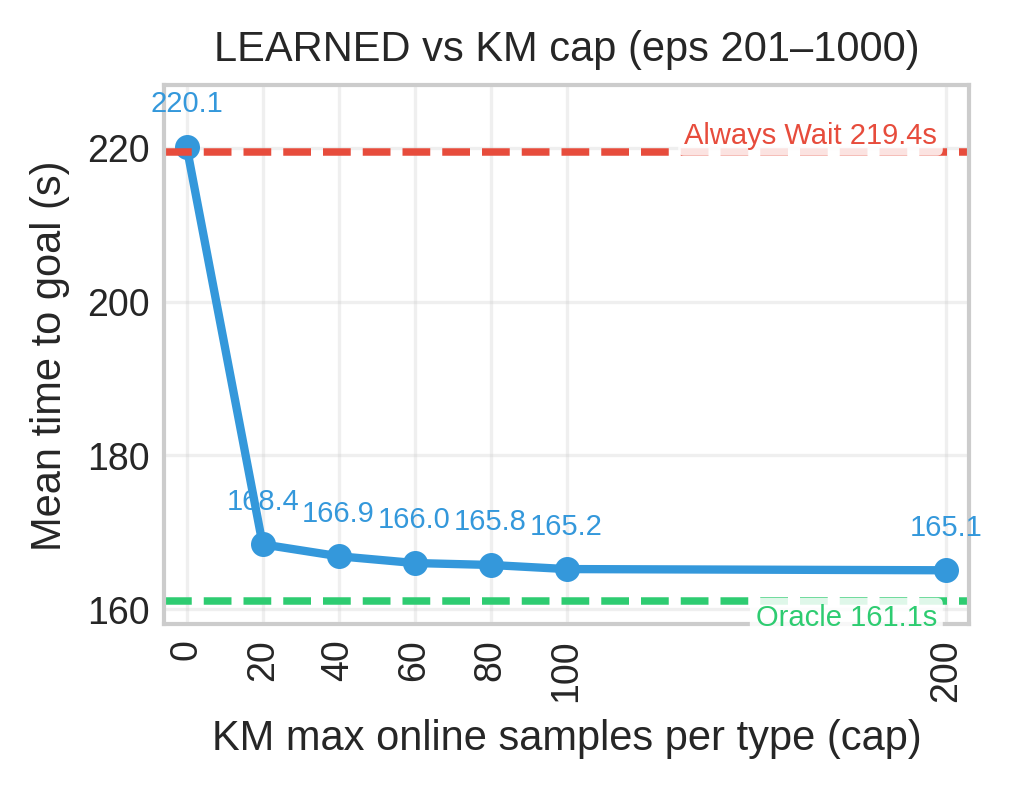}
        \caption{Effect of limiting Kaplan--Meier samples.}
        \label{fig:km_data_cap}
    \end{subfigure}
    \captionsetup{font=footnotesize}
    \caption{Simulation performance and data-efficiency results. 
    (a) The learned policy improves with experience, outperforms the heuristic baselines, and approaches the oracle policy. 
    (b) Performance improves sharply with a small number of clearance samples and then saturates near the oracle policy.}
    \label{fig:simulation_results_combined}
    \vspace{-0.8em}
\end{figure}

For each method, we run repeated navigation episodes on the same graph and obstacle process and record relevant statistics. Table~\ref{tab:sim_results} summarizes 1000 episodes averaged over 100 random seeds. Figure~\ref{fig:sim_results} shows OSCAR improving rapidly toward oracle performance and outperforming heuristic baselines. The no-memory ablation falls between the heuristic baselines and the full learned method, isolating the benefit of survival learning while showing that obstacle memory is needed for near-oracle performance.
Always waiting wastes time at persistent blockages, while immediate rerouting incurs unnecessary detours. The rule-based baseline cannot adapt to the actual persistence statistics of the environment. Greedy CTP performs especially poorly because it permanently removes each observed blocked edge for the remainder of the episode. As edges are removed, the planner can become trapped with no available path to the goal; these failed episodes are assigned the \(1\,\mathrm{hr}\) timeout as their time-to-goal. 

\begin{figure}[htbp]
\vspace{-0.8em}
    \centering

    \begin{minipage}{0.49\linewidth}
        \centering
        \includegraphics[width=\linewidth]{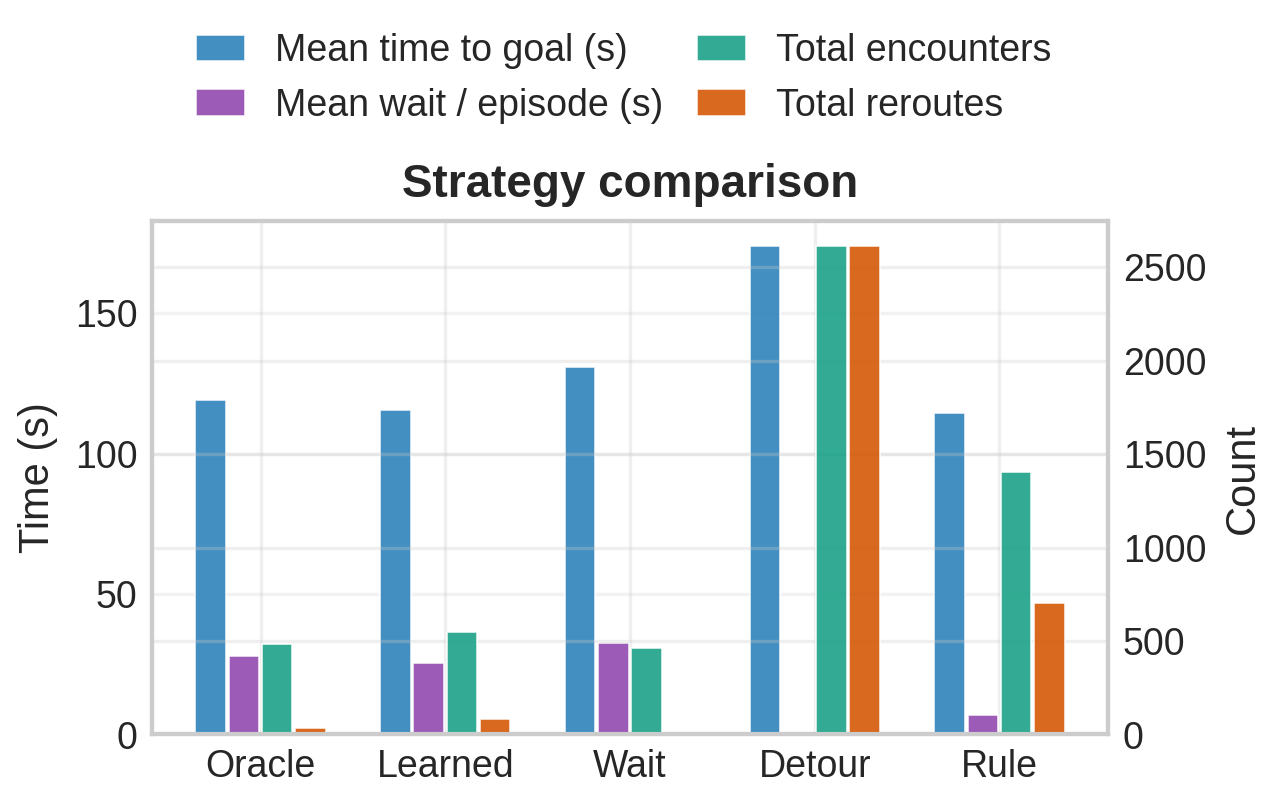}
        \vspace{-0.5em}
        \centerline{\small (a) Strategy comparison}
    \end{minipage}
    \hfill
    \begin{minipage}{0.49\linewidth}
        \centering
        \includegraphics[width=\linewidth]{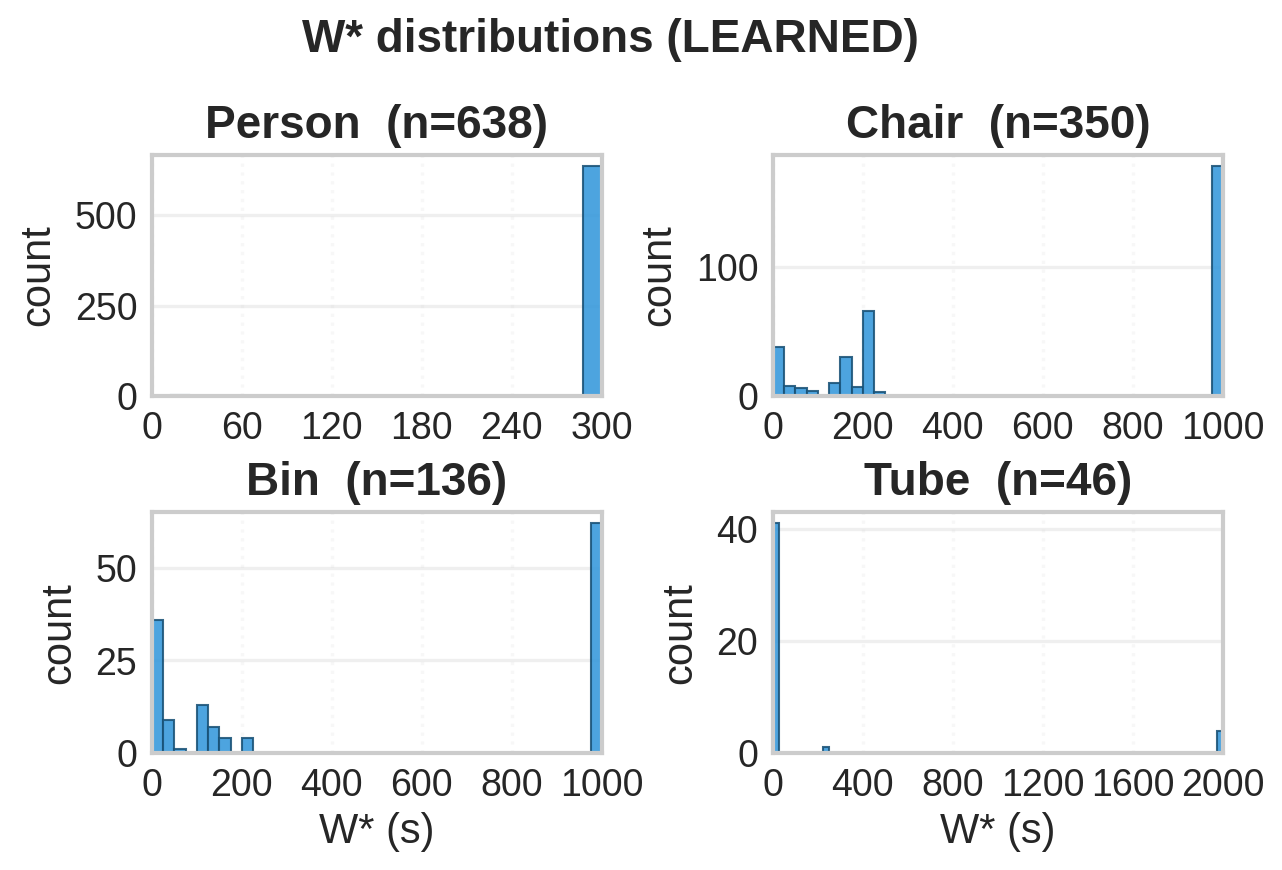}
        \vspace{-0.5em}
        \centerline{\small (b) Learned patience thresholds}
    \end{minipage}
    \captionsetup{font=footnotesize}
    \caption{
    Simulation analysis of the learned KM policy on a grid graph.
    (a) Comparison of navigation strategies across 300 episodes,
    showing mean time-to-goal, waiting time, obstacle encounters,
    and reroutes.
    (b) Distribution of selected patience thresholds $W^\star$
    for each obstacle class. OSCAR generally waits for people,
    balances waiting and rerouting for chairs and bins, and
    reroutes immediately for tubes.
    }
    \label{fig:sim_strategy_wstar}
    \vspace{-1em}
\end{figure}

We also evaluate data efficiency by capping Kaplan--Meier samples per obstacle class and reporting mean time-to-goal after convergence (200 episodes). Figure~\ref{fig:km_data_cap} shows that performance improves sharply with a small number of samples before saturating near the oracle, indicating that extensive clearance data are unnecessary. We further examine the learned policy's decisions on a grid-based graph where the rule-based baseline is expected to perform well (Fig.~\ref{fig:sim_strategy_wstar}). OSCAR remains competitive while producing meaningful patience thresholds: it waits for people, often waits for chairs and bins to avoid risky detours, and reroutes immediately for tubes. Additional experiments demonstrate that OSCAR generalizes across graph topologies and blockage probabilities (\ref{app:generalization}).

\begin{table}[htbp]
    \centering
    \captionsetup{font=footnotesize}
    \caption{
    Summary of simulation results over 100 runs (1000 episodes per run).
    All metrics are averaged per episode. Reroutes denotes the number
    of route changes, waiting time is the cumulative time spent waiting,
    and blocked edges counts obstacle encounters, including repeated
    encounters with the same edge. OSCAR achieves the lowest mean
    time-to-goal among non-oracle methods by balancing waiting and rerouting.
    }
    \label{tab:sim_results}
    \resizebox{\columnwidth}{!}{%
    \begin{tabular}{lccccc}
        \toprule
        Method & Time-to-goal (s) $\downarrow$ & Success rate $\uparrow$ & Reroutes & Waiting time & Blocked edges \\
        \midrule
            Always Wait & 218.2 & 100.0\% & 0.0 & 186.2 & 2.8 \\
            Always Reroute & 264.7 & 100.0\% & 24.2 & 0.0 & 24.2 \\
            Rule-Based & 210.5 & 100.0\% & 11.9 & 29.0 & 16.1 \\
            Greedy CTP & 3135.7 & 13.1\% & 3.4 & 0.0 & 2.5 \\
            
            OSCAR w/o Memory & 185.9 & 100.0\% & 1.8 & 149.8 & 4.5 \\
            OSCAR (Ours) & 160.7 & 100.0\% & 0.5 & 121.9 & 3.4 \\
            \midrule 
            Oracle & 159.3 & 100.0\% & 0.5 & 121.7 & 3.3 \\
        \bottomrule
    \end{tabular}
    }
\end{table}

\noindent
\begin{minipage}[t]{0.68\linewidth}
\vspace{0pt}
\textbf{Right-censoring ablation.}
To isolate the contribution of censored observations, we compare OSCAR against an otherwise identical variant that discards observations when the robot reroutes before obstacle clearance. We evaluate both methods in a cluttered environment with 0.3\,m edges and bimodal clearance times (5\,s and 250\,s), averaging over 10 random seeds.
As shown in Fig.~\ref{fig:censor_ablation}, incorporating censored observations improves navigation efficiency during online learning and allows OSCAR to approach oracle performance. OSCAR estimates the expected residual clearance time at 241.4\,s, close to the ground-truth value of 246.1\,s, whereas discarding censored observations yields 205.8\,s, a 16\% underestimate that leads to less effective wait-or-reroute decisions.
\end{minipage}\hfill
\begin{minipage}[t]{0.28\linewidth}
\vspace{0pt}
\centering
\includegraphics[width=\linewidth]
{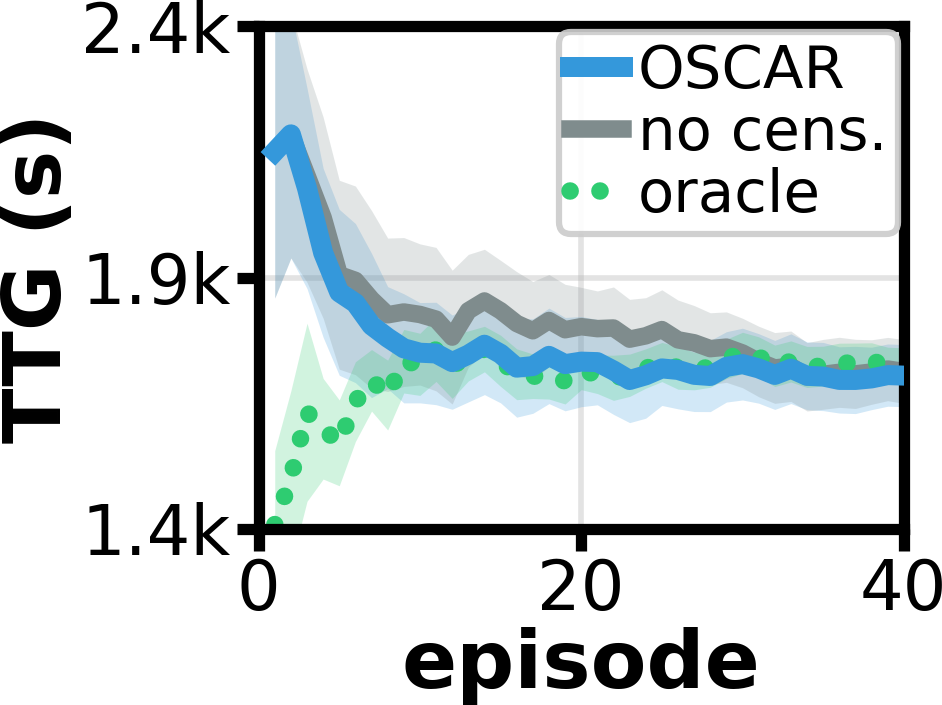}
\vspace{-0.5em}
\captionsetup{font=footnotesize,skip=2pt}
\captionof{figure}{
Navigation performance with and without
censored observations (10 seeds).
}
\label{fig:censor_ablation}
\end{minipage}

\noindent
\begin{minipage}[t]{0.68\linewidth}
\vspace{0pt}
\textbf{Computational scalability.}
We evaluate the computational cost of selecting $W^\star$ on lattice graphs with 49--1024 vertices and 4, 8, or 16 obstacle classes (Fig.~\ref{fig:scalability}). Computation time grows approximately linearly with graph size and exhibits limited sensitivity to the number of obstacle classes. Even at 1024 vertices, mean and 95th-percentile computation times remain below 0.172\,s and 0.276\,s, respectively. Capping the number of candidate clearance times also prevents computation time from growing indefinitely as observations accumulate.
\end{minipage}\hfill
\begin{minipage}[t]{0.28\linewidth}
\vspace{0pt}
\centering
\includegraphics[width=0.95\linewidth]
{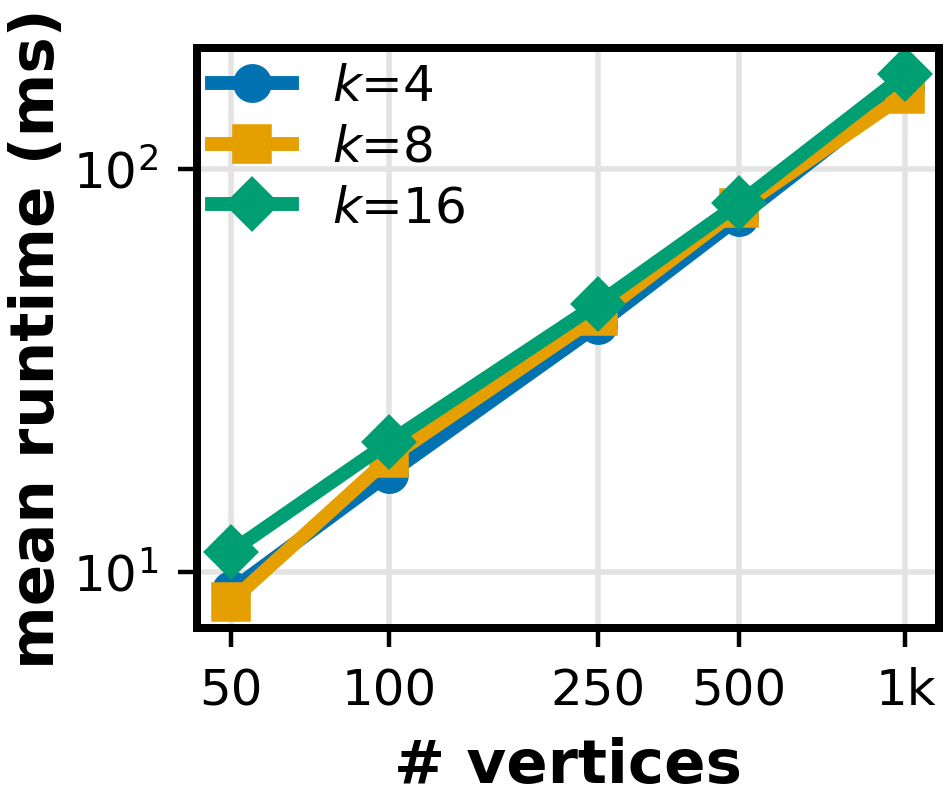}
\captionsetup{font=footnotesize,skip=2pt}
\captionof{figure}{
Mean $W^\star$ computation time by graph size
and obstacle class count.
}
\label{fig:scalability}
\vspace{-0.7em}
\end{minipage}

\vspace{-0.8em}
\subsection{Real-World Experiments}
\label{realworldexp}
\begin{figure}[htbp]
    \centering
    \begin{minipage}[t]{0.34\linewidth}
        \centering
        \includegraphics[width=\linewidth, height=0.82\linewidth]{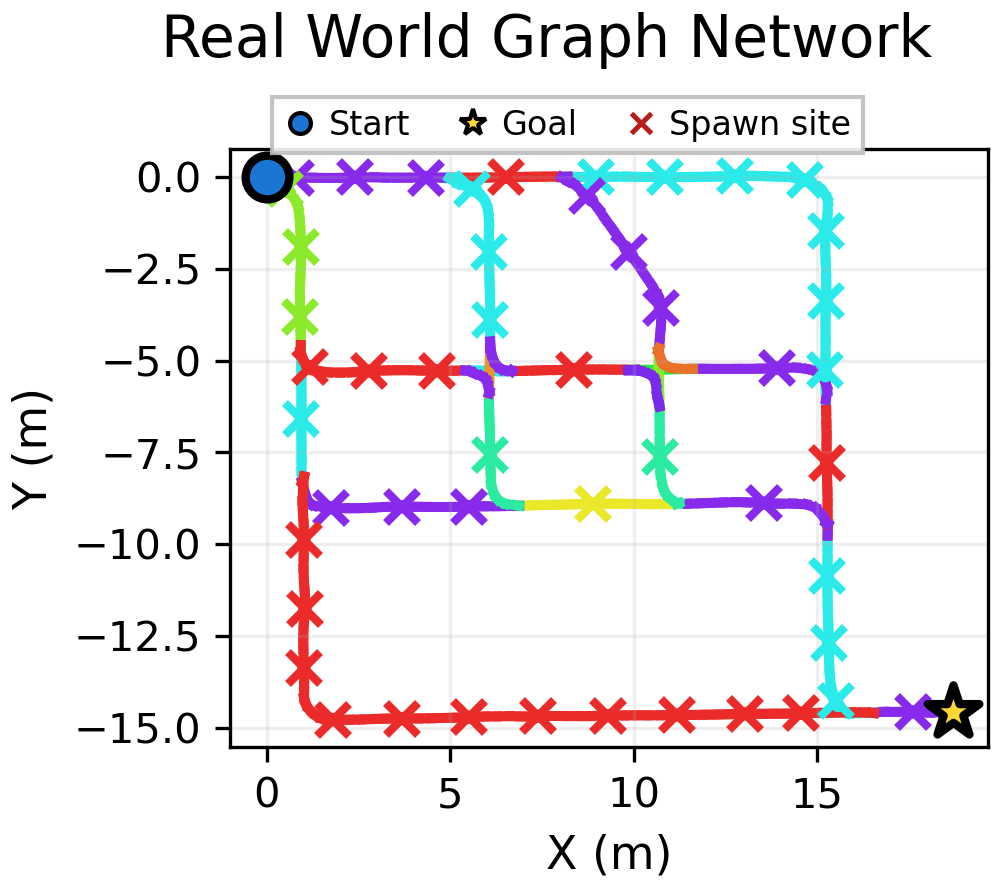}
        \centerline{\small (a) Atrium graph}
    \end{minipage}
    \hfill
    \begin{minipage}[t]{0.62\linewidth}
        \centering
        \includegraphics[width=\linewidth, height=0.45\linewidth]{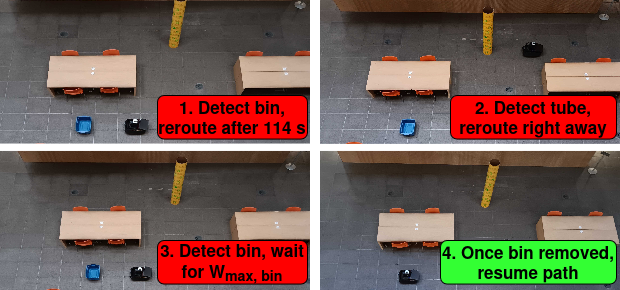}
        \centerline{\small (b) Example obstacle encounter sequence}
    \end{minipage}
    \captionsetup{font=footnotesize}
    \caption{
    Real-world deployment setup.
    (a) Graph and allowable obstacle spawn locations
    in the atrium. Each color represents a distinct
    corridor segment connecting junctions in the
    taught-route network.
    (b) Representative obstacle encounter sequence
    illustrating wait and reroute decisions.
    }
    \label{fig:real_world_setup}
    \vspace{-1em}
\end{figure}

We evaluate the method on a physical robot in a university atrium. The obstacle classes are person, chair, bin, and tube. Figure~\ref{fig:real_world_setup} shows the atrium graph, obstacle spawning locations, and a representative obstacle encounter sequence. We use the teach-and-repeat~\citep{ltr} navigation system: in the teach phase, we record a graph network with multiple branches, and during the repeat phase the robot autonomously navigates this graph using our policy. When the robot detects a blockage, it extracts the LiDAR reflectivity image, generates an obstacle mask, and queries ChatGPT~\citep{openai2026gpt53} with the mask and obstacle-class labels to select the corresponding survival model. 
The prompt to ChatGPT is shown in ~\ref{app:vlm_prompt}. Unorchestrated results in a person/chair-only setting appear in ~\ref{app:unorchestrated_person_chair}. A supplementary video is available at \url{https://youtu.be/h4LEMJS2tpQ}.

\begin{figure*}[h]
    \centering

    \begin{minipage}{0.32\linewidth}
        \centering
        \includegraphics[width=\linewidth]{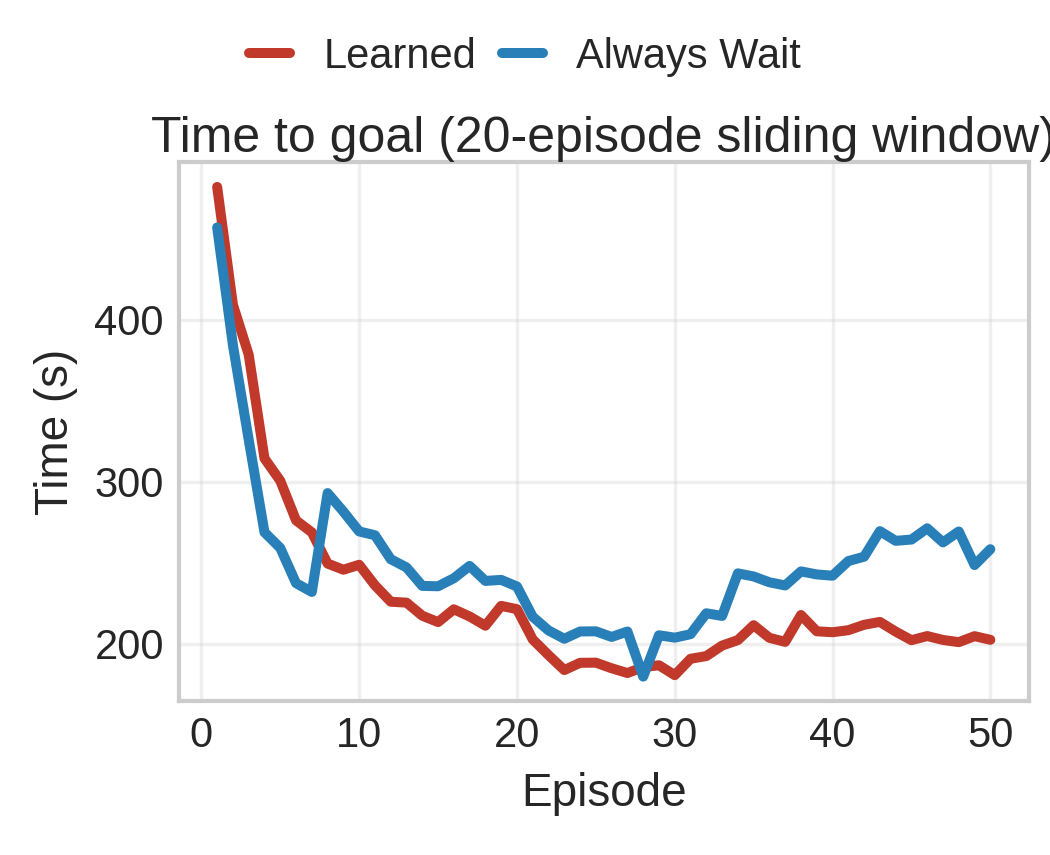}
        \vspace{-0.5em}
        \centerline{\small (a) Time-to-goal}
    \end{minipage}
    \hfill
    \begin{minipage}{0.32\linewidth}
        \centering
        \includegraphics[width=\linewidth]{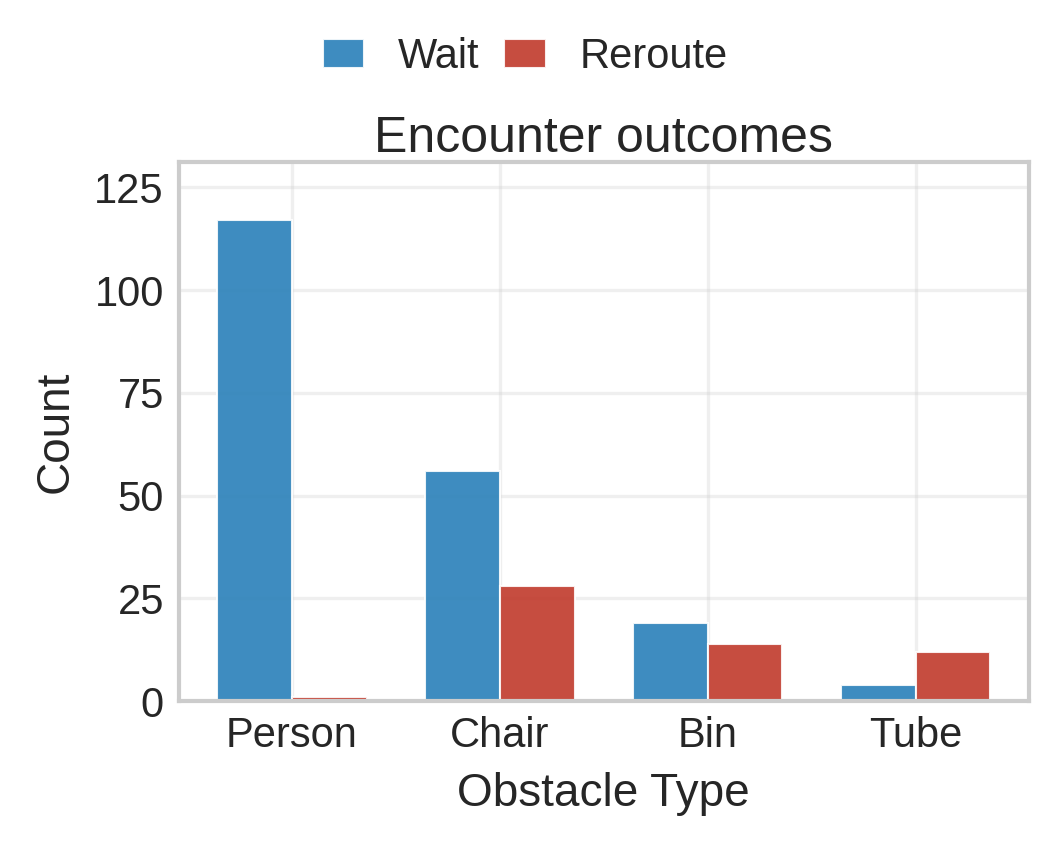}
        \vspace{-0.5em}
        \centerline{\small (b) Wait vs. reroute decisions}
    \end{minipage}
    \hfill
    \begin{minipage}{0.32\linewidth}
        \centering
        \includegraphics[width=\linewidth]{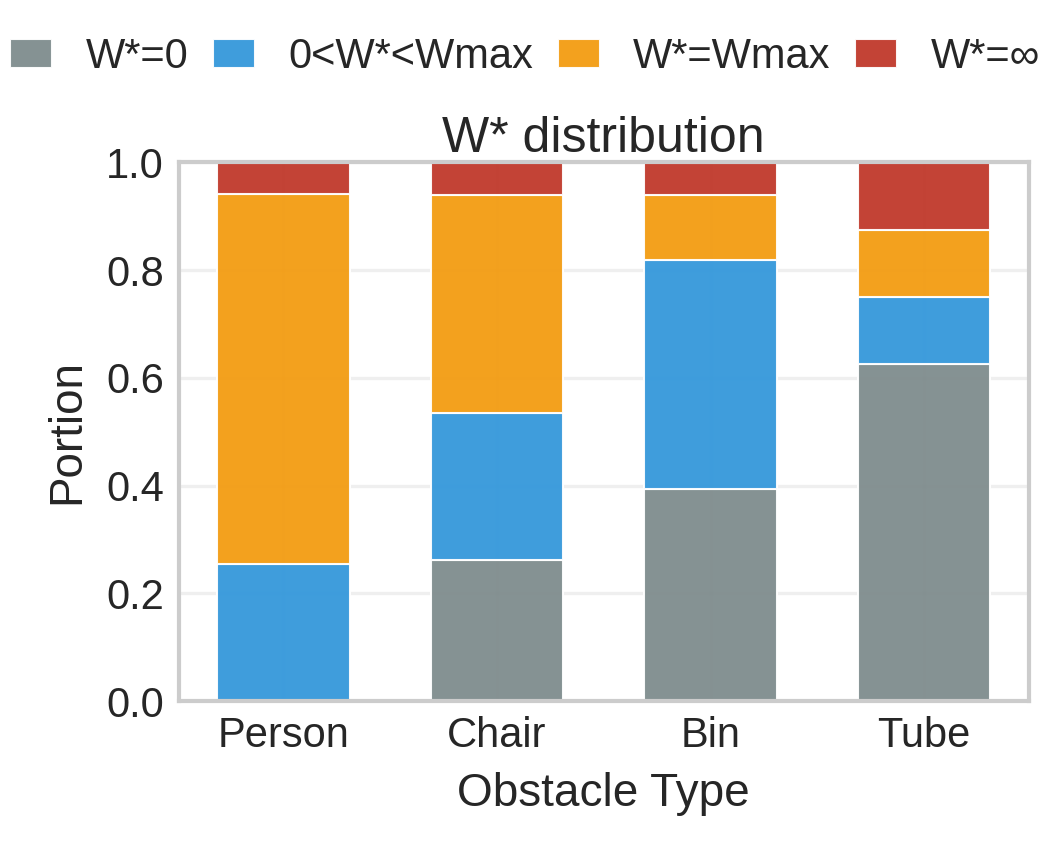}
        \vspace{-0.5em}
        \centerline{\small (c) Selected patience thresholds}
    \end{minipage}

    \captionsetup{font=footnotesize}
    \caption{Real-world experiment results. 
    (a) Time-to-goal over 50 navigation episodes, showing how performance improves with online experience. 
    (b) Frequency of wait and reroute decisions made by the learned policy during deployment. 
    (c) Distribution of selected patience thresholds $W^\star$.}
    \label{fig:real_world_results}
    \vspace{-0.8em}
\end{figure*}

We navigate inside an atrium for 50 episodes, creating dynamic scenes using our Poisson spawning process and ground-truth clearance-time distributions. For every episode, we record time-to-goal, and the learned method updates its survival estimates online from the resulting censored or uncensored observations. Figure~\ref{fig:real_world_results} summarizes the real-world deployment results. In Fig.~\ref{fig:real_world_results}(a), the time-to-goal decreases over repeated episodes, outperforms the always-wait baseline, and converges to approximately \(200\,\mathrm{s}\), indicating that the learned survival estimates improve with experience. Figure~\ref{fig:real_world_results}(b) shows that the policy waits for people, uses a more balanced wait-reroute strategy for chairs and bins, and mostly reroutes for tubes.
This behavior is reflected in Fig.~\ref{fig:real_world_results}(c): people are assigned \(W^\star=W_{\max}\), chairs and bins receive a mixture of immediate reroutes, finite nonzero waits, and long waits, while tubes are often assigned \(W^\star=0\). We set the class-specific \(W_{\max}\) values empirically by inspecting \(J(W)\) curves and choosing a horizon after which the objective was effectively flat. Cases with \(W^\star=\infty\) correspond to encounters where no alternate path is available.



\section{Limitations and Future Work}
\label{sec:limitations}
\textbf{Decision Module:} First, the current wait-or-reroute decision estimates the cost of \emph{committing} to a different path, not accounting for the fact that the robot may later encounter another obstacle and replan again. As a result, the cost of rerouting can be overestimated, causing the policy to reroute less often than desired. Second, the KM estimator assumes independent censoring, meaning that the censoring time does not depend on the unobserved clearance time~\citep{klein2003survival}, conditional on the modeled variables. In our simulator, obstacle clearance times are sampled independently given their class, while the patience threshold $W^\star$ is determined from previously collected observations and the current planning problem. Since the threshold is selected without information about the current obstacle's unobserved clearance time, adaptive censoring does not itself violate the independence assumption under our simulation assumptions. In real-world environments, however, unmodeled contextual factors may influence both clearance times and the patience thresholds selected through the planning objective. For example, people may linger longer in certain locations where inexpensive alternative routes also encourage shorter waiting times. In practice, the simulation results in Fig.~\ref{fig:sim_results} show that the learned policy still converges near the oracle. However, a more rigorous treatment of outcome-adaptive censoring, or theoretical guarantees for this feedback setting, is an important direction for future work.

\textbf{Environment and perception assumptions:} The planner uses a single learned blockage probability for each obstacle type shared across all edges, whereas some edges are more likely to be blocked than others due to spatial layout, human activity, etc. Each encounter is also treated as a new obstacle observation, even when a later observation may correspond to the same physical obstacle; distinguishing persistent obstacles from new ones would improve the survival estimates. Finally, our experiments assume reliable obstacle classification, which was feasible given the small, visually distinct obstacle set. Our survival models currently condition on discrete obstacle classes, but the framework could accommodate richer contextual information. For example, a VLM could distinguish between a walking person, a seated person, and a chair with a nearby person, allowing separate clearance-time distributions for each category. Beyond classification, a VLM could provide context-dependent prior estimates of obstacle clearance times; these predictions could then be refined using online clearance and right-censored observations, combining prior semantic knowledge with experience-based survival learning.

\section{Conclusion}
\label{sec:conclusion}
We presented OSCAR, combining online obstacle-survival learning,
obstacle memory, and time-dependent graph planning to decide when
to wait or reroute. In simulation, OSCAR approaches oracle
performance and outperforms heuristic baselines, with ablations
demonstrating the benefits of right-censored observations and
obstacle memory. Scalability experiments demonstrate efficient
planning on larger graphs, while 50 real-world navigation episodes
confirm that OSCAR adapts its patience thresholds to different
obstacle classes and blockage scenarios as experience accumulates.
\clearpage

\bibliography{example}  

\clearpage
\appendix
\renewcommand{\thesection}{Appendix \Alph{section}}

\section{Functional Time-Dependent Planner and Proofs}
\label{app:functional_planner}

This appendix gives the functional planner used to compute
\(A_{\mathrm{clear}}\) and \(A_{\mathrm{avoid}}\), and proves the properties claimed
in Sec.~\ref{sec:method}. Throughout, times are measured relative to the
blocked-edge encounter time \(t_0\). Let \(v_0\) denote the vertex at which the
robot encounters the blocked edge \(e_b\), and let \(g\) denote the goal.

\subsection{FIFO Edge Arrival Functions}

For an edge \(e=(u,v)\), define the edge-arrival function
\[
\Phi_e(t)=t+w(e)+\widehat{\Delta}(e,t),
\]
where \(t\) is the planned arrival time at the tail \(u\). The edge is FIFO if
\[
t_1\le t_2
\quad\Longrightarrow\quad
\Phi_e(t_1)\le \Phi_e(t_2).
\]
Equivalently, departing later from the tail of an edge cannot lead to arriving
earlier at its head.

If edge \(e\) has no remembered obstacle state, then
\[
\widehat{\Delta}(e,t)=\widehat{\Delta}_{\mathrm{new}},
\]
so
\[
\Phi_e(t)=t+w(e)+\widehat{\Delta}_{\mathrm{new}},
\qquad
\frac{d\Phi_e}{dt}=1\ge 0.
\]
Thus, edges without remembered obstacle state are FIFO.

Now consider an edge with remembered obstacle state
\(M(e)=(k_e,t_{\mathrm{first}}(e),t_{\mathrm{last}}(e))\). Let
\[
a_e=t_{\mathrm{last}}(e)-t_{\mathrm{first}}(e),
\qquad
b_e(t)=t-t_{\mathrm{first}}(e),
\]
where \(a_e\) is the obstacle age when it was last confirmed present and
\(b_e(t)\) is the obstacle age at planned arrival time \(t\). For readability,
write \(S(\cdot)=S_{k_e}(\cdot)\). The remembered-obstacle delay term is
\[
\widehat{\Delta}(e,t)
=
q_e(t)\widehat{\mu}^{\mathrm{old}}_e(t)
+
(1-q_e(t))\widehat{\Delta}_{\mathrm{new}},
\]
where
\[
q_e(t)=\frac{S(b_e(t))}{S(a_e)}
\]
and
\[
\widehat{\mu}^{\mathrm{old}}_e(t)
=
\int_0^\infty
\frac{S(b_e(t)+r)}{S(b_e(t))}\,dr.
\]
Multiplying the first two terms gives
\[
q_e(t)\widehat{\mu}^{\mathrm{old}}_e(t)
=
\frac{1}{S(a_e)}
\int_0^\infty S(b_e(t)+r)\,dr
=
\frac{1}{S(a_e)}
\int_{b_e(t)}^\infty S(r)\,dr.
\]
Therefore,
\[
\Phi_e(t)
=
t+w(e)
+
\frac{1}{S(a_e)}
\int_{b_e(t)}^\infty S(r)\,dr
+
\left(1-\frac{S(b_e(t))}{S(a_e)}\right)
\widehat{\Delta}_{\mathrm{new}}.
\]

Since \(db_e(t)/dt=1\), differentiating gives
\[
\frac{d\Phi_e}{dt}
=
1
-
\frac{S(b_e(t))}{S(a_e)}
-
\frac{S'(b_e(t))}{S(a_e)}
\widehat{\Delta}_{\mathrm{new}}.
\]
Because \(S\) is a survival function, it is non-increasing, so
\[
S'(b_e(t))\le 0.
\]
Also, since \(b_e(t)\ge a_e\) for planned arrivals after the last confirmed
observation and \(S\) is non-increasing,
\[
S(b_e(t))\le S(a_e).
\]
Hence,
\[
1-\frac{S(b_e(t))}{S(a_e)}\ge 0,
\]
and
\[
-\frac{S'(b_e(t))}{S(a_e)}
\widehat{\Delta}_{\mathrm{new}}
\ge 0.
\]
Therefore,
\[
\frac{d\Phi_e}{dt}\ge 0.
\]
Thus, \(\Phi_e(t)\) is non-decreasing, and the remembered-obstacle edge
relaxation is FIFO.

For the Kaplan--Meier estimator, \(S\) is piecewise constant rather than
everywhere differentiable. The same argument holds piecewise: between KM event
times, \(S'(t)=0\), so
\[
\frac{d\Phi_e}{dt}
=
1-\frac{S(b_e(t))}{S(a_e)}
\ge 0.
\]
At a KM event time, \(S\) jumps downward. The integral term remains continuous,
while
\[
\left(1-\frac{S(b_e(t))}{S(a_e)}\right)
\widehat{\Delta}_{\mathrm{new}}
\]
jumps upward or stays constant. Therefore, \(\Phi_e\) has nonnegative slope
between breakpoints and no downward jumps at breakpoints. Hence, the KM edge
arrival function is also non-decreasing and FIFO.

\subsection{Functional Bellman--Ford Computation}

We compute arrival-time functions by propagating functions instead of scalar
arrival times. Let \(B_v(z)\) denote the earliest arrival time at vertex \(v\),
measured relative to \(t_0\), as a function of the scalar parameter \(z\). The
parameter \(z\) is either the clearance time \(c\) for \(A_{\mathrm{clear}}\), or
the patience threshold \(W\) for \(A_{\mathrm{avoid}}\).

For a candidate function \(B_u(z)\) at the tail of edge \(e=(u,v)\), the
functional edge relaxation is
\[
B_v(z)
\leftarrow
\min\left\{
B_v(z),\,
\Phi_e(B_u(z))
\right\}.
\]
The minimum is taken pointwise over the function domain. Since \(\Phi_e\) is
FIFO and \(B_u(z)\) is non-decreasing, the composed function
\(\Phi_e(B_u(z))\) is non-decreasing. A pointwise minimum of non-decreasing
functions is also non-decreasing. Therefore, every Bellman--Ford relaxation
preserves monotonicity.

With Kaplan--Meier survival estimates, the functions involved are
piecewise-affine, with breakpoints at clearance-time observations and at
breakpoints introduced by edge relaxations. Thus, the same Bellman--Ford
procedure can be applied directly to function representations. In practice, the
learned implementation evaluates the finite KM candidate set using
time-dependent Dijkstra, while the functional Bellman--Ford view gives the
underlying continuous-time construction.

\paragraph{Clear case.}
For \(A_{\mathrm{clear}}(c)\), the obstacle on \(e_b\) clears after elapsed time
\(c\). The robot is still at \(v_0\) at time \(c\), so the initial function is
\[
B^{\mathrm{clear}}_{v_0}(c)=c,
\]
and all other vertices are initialized as
\[
B^{\mathrm{clear}}_{v}(c)=+\infty,
\qquad v\ne v_0.
\]
The blocked edge \(e_b\) is treated as available after time \(c\), since the
condition defining \(A_{\mathrm{clear}}(c)\) is that the encountered obstacle has
cleared. Functional Bellman--Ford then performs the relaxation
\[
B^{\mathrm{clear}}_{v}(c)
\leftarrow
\min\left\{
B^{\mathrm{clear}}_{v}(c),\,
\Phi_e(B^{\mathrm{clear}}_{u}(c))
\right\}
\]
for each edge \(e=(u,v)\), repeated for at most \(|V|-1\) passes. The resulting
clearance-conditioned time-to-goal is
\[
A_{\mathrm{clear}}(c)=B^{\mathrm{clear}}_{g}(c).
\]

\paragraph{Avoid case.}
For \(A_{\mathrm{avoid}}(W)\), the robot waits until elapsed time \(W\), the
obstacle has not cleared, and then replans with \(e_b\) forbidden. The initial
function is
\[
B^{\mathrm{avoid}}_{v_0}(W)=W,
\]
and all other vertices are initialized as
\[
B^{\mathrm{avoid}}_{v}(W)=+\infty,
\qquad v\ne v_0.
\]
The functional Bellman--Ford relaxation is then applied on the graph
\((V,E\setminus\{e_b\})\):
\[
B^{\mathrm{avoid}}_{v}(W)
\leftarrow
\min\left\{
B^{\mathrm{avoid}}_{v}(W),\,
\Phi_e(B^{\mathrm{avoid}}_{u}(W))
\right\},
\qquad e=(u,v)\ne e_b.
\]
After at most \(|V|-1\) passes, the avoid-conditioned time-to-goal is
\[
A_{\mathrm{avoid}}(W)=B^{\mathrm{avoid}}_{g}(W).
\]

\subsection{Monotonicity of \(A_{\mathrm{clear}}\) and \(A_{\mathrm{avoid}}\)}

We prove that both \(A_{\mathrm{clear}}\) and \(A_{\mathrm{avoid}}\) are
non-decreasing in their arguments.

First consider \(A_{\mathrm{clear}}\). At the start vertex \(v_0\),
\[
B^{\mathrm{clear}}_{v_0}(c)=c,
\]
which is non-decreasing in \(c\). All other initial functions are \(+\infty\),
which are also non-decreasing. Each edge relaxation composes the current
tail-arrival function with a FIFO edge-arrival function and then takes a
pointwise minimum with the previous function. As shown above, both operations
preserve non-decreasing functions. Therefore, after all Bellman--Ford passes,
\(B^{\mathrm{clear}}_g(c)\) is non-decreasing. Hence,
\[
A_{\mathrm{clear}}(c)=B^{\mathrm{clear}}_g(c)
\]
is non-decreasing.

The same argument applies to \(A_{\mathrm{avoid}}\). The initialization
\[
B^{\mathrm{avoid}}_{v_0}(W)=W
\]
is non-decreasing in \(W\), and all other vertices are initialized to
\(+\infty\). The relaxations are performed over \(E\setminus\{e_b\}\), but each
remaining edge is still FIFO. Therefore, every relaxation preserves
non-decreasing functions, and the final goal-arrival function
\(B^{\mathrm{avoid}}_g(W)\) is non-decreasing. Hence,
\[
A_{\mathrm{avoid}}(W)=B^{\mathrm{avoid}}_g(W)
\]
is non-decreasing.

Intuitively, both quantities are measured from the original encounter time
\(t_0\). If the robot waits longer before the obstacle clears, then
\(A_{\mathrm{clear}}(c)=c+x(c)\), where \(x(c)\) is the additional time needed to
reach the goal after clearance. Similarly, if the robot waits longer before
giving up, then \(A_{\mathrm{avoid}}(W)=W+x(W)\). Although the downstream
remaining travel time \(x(\cdot)\) may decrease because later arrival can reduce
expected obstacle delay elsewhere, FIFO guarantees that this decrease cannot
more than cancel the additional elapsed waiting time.

\subsection{Finite Candidate Set for the KM Objective}

For a fixed obstacle class \(k\), let the fitted Kaplan--Meier distribution have
event times
\[
0<\tau_1<\tau_2<\cdots<\tau_J.
\]
The learned wait-or-reroute objective is
\[
\widehat{J}(W)
=
\sum_{\tau_j\le W}
\widehat{p}_{k,j} A_{\mathrm{clear}}(\tau_j)
+
\widehat{S}_k(W)A_{\mathrm{avoid}}(W),
\]
where
\[
\widehat{p}_{k,j}
=
\widehat{S}_k(\tau_{j-1})-\widehat{S}_k(\tau_j).
\]

Between consecutive KM event times, the survival estimate is constant and no
new clearance probability mass is added to the summation. Therefore, for any
interval \([\tau_j,\tau_{j+1})\), the first term in \(\widehat{J}(W)\) is
constant and \(\widehat{S}_k(W)\) is constant. On that interval,
\[
\widehat{J}(W)
=
C_j
+
\widehat{S}_k(\tau_j) A_{\mathrm{avoid}}(W),
\]
for some constant \(C_j\). Since \(A_{\mathrm{avoid}}(W)\) is non-decreasing and
\(\widehat{S}_k(\tau_j)\ge 0\), \(\widehat{J}(W)\) is non-decreasing on
\([\tau_j,\tau_{j+1})\). Hence, the minimum over that interval occurs at its
left endpoint, \(\tau_j\). The same argument applies to \([0,\tau_1)\), where
the minimum occurs at \(W=0\).

It follows that the global minimizer over a bounded horizon
\([0,W_{\max}]\) is attained at a finite candidate time:
\[
W^\star
=
\operatorname*{arg\,min}_{W\in\mathcal{C}}
\widehat{J}(W),
\qquad
\mathcal{C}
=
\{0\}
\cup
\{\tau_j:\tau_j\le W_{\max}\}.
\]
Including \(W_{\max}\) in the implementation is harmless and convenient:
\[
\mathcal{C}_{\mathrm{impl}}
=
\{0,W_{\max}\}
\cup
\{\tau_j:\tau_j\le W_{\max}\}.
\]
Thus, under the KM survival estimate, the patience-threshold search reduces to
evaluating \(\widehat{J}(W)\) on a finite set of survival event times, together
with the horizon endpoints.
\newpage
\section{Simulation Details}
\label{app:simulation_details}

\begin{figure}[h]
    \centering
    \includegraphics[width=0.6\linewidth]{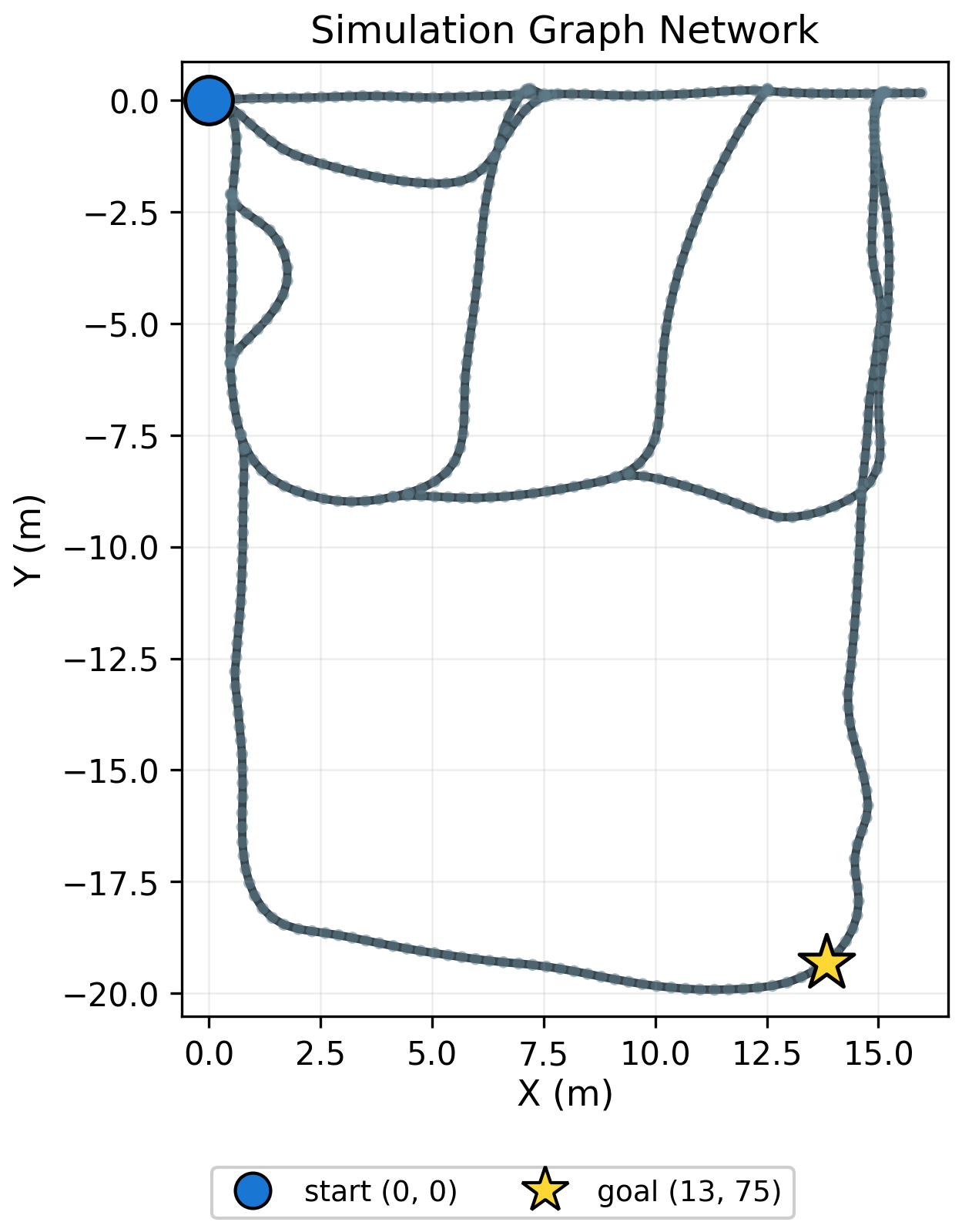}
    \caption{Graph used for simulation experiments. The graph contains multiple
    alternate routes, allowing policies to trade off waiting against rerouting
    when obstacles temporarily block edges.}
    \label{fig:sim_graph}
\end{figure}

This appendix describes the simulation setup used for the results in
Sec.~\ref{sec:result}. The simulator operates on a graph-based route network
whose vertices represent robot poses and whose edges represent feasible motions.
The graph contains multiple alternate routes between the start and goal, allowing
the policies to make meaningful wait-or-reroute decisions when an edge becomes
blocked. The robot travels along graph edges at a fixed speed of
\(0.95\,\mathrm{m/s}\). The start vertex is fixed, and the goal vertex is
selected as the vertex farthest from the start in the graph. A visualization of
the simulation graph is shown in Fig.~\ref{fig:sim_graph}.

\subsection{Ground-Truth Obstacle Distributions}

Each obstacle class is assigned a ground-truth lifetime distribution. Let
\(C^{(k)}\) denote the clearance time of an obstacle of class \(k\), measured
from the time the obstacle is spawned. We use lognormal distributions,
\[
C^{(k)} \sim \mathrm{LogNormal}(\mu_k,\sigma_k),
\]
with parameters chosen to produce different obstacle-persistence regimes.
People clear quickly, chairs and bins persist for intermediate durations, and
tubes persist the longest. The simulator uses four obstacle classes:
person, chair, bin, and tube.

\begin{figure}[h]
    \centering
    \includegraphics[width=\linewidth]{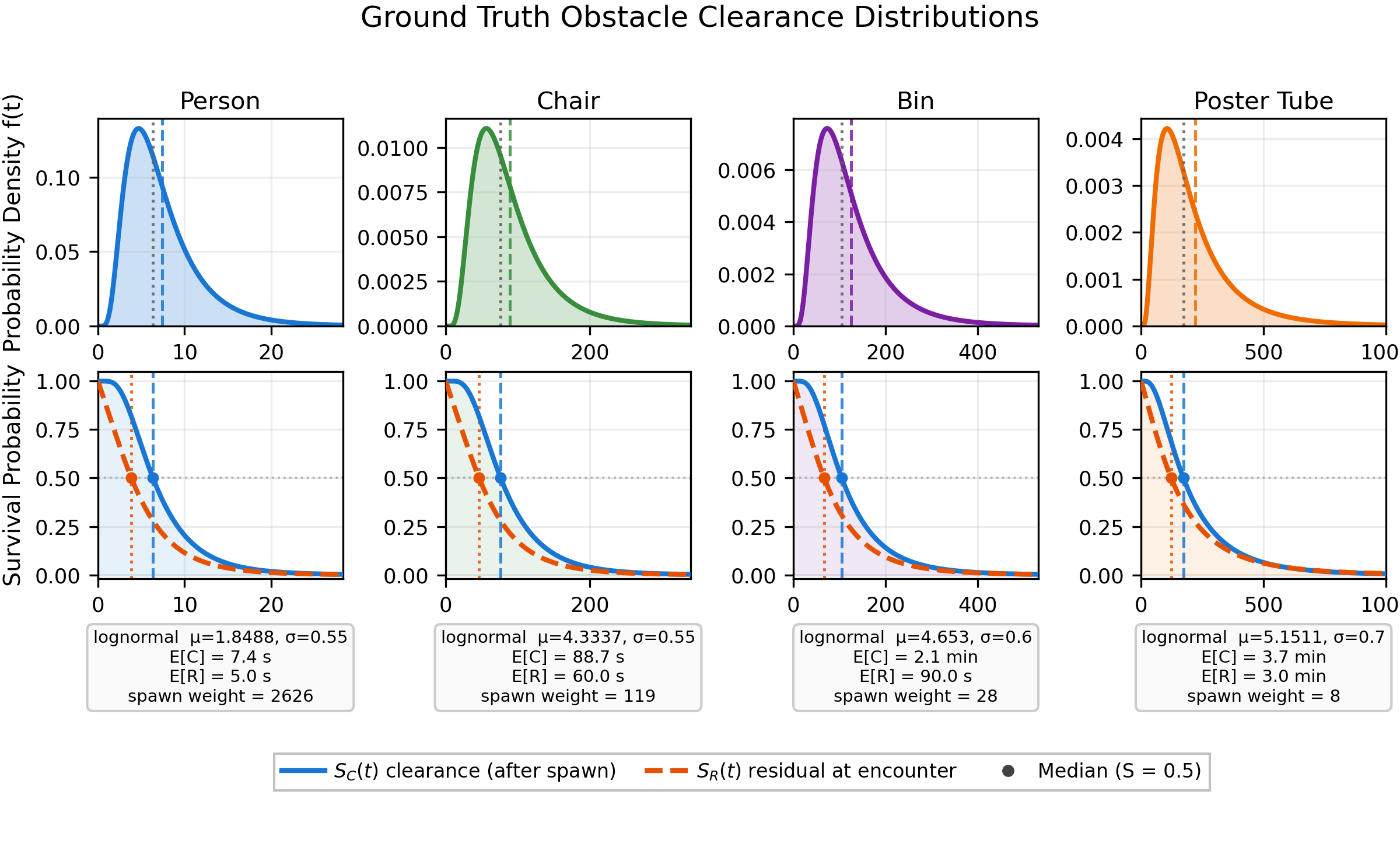}
    \caption{Ground-truth obstacle clearance distributions used in simulation.
    Top: clearance-time densities from obstacle spawn time. Bottom: clearance
    survival \(S_C(t)\) and residual survival \(S_R(t)\) at obstacle encounter.
    The residual distribution is the distribution relevant to the robot's
    wait-or-reroute decision.}
    \label{fig:sim_ground_truth_distributions}
\end{figure}

Figure~\ref{fig:sim_ground_truth_distributions} shows the ground-truth obstacle
distributions used in simulation. The top row shows the clearance-time density
\(f_C(t)\) for each obstacle class, where clearance time is measured from the
obstacle spawn time. The bottom row compares the clearance survival function
\[
S_C^{(k)}(t)=\mathbb{P}(C^{(k)}>t)
\]
with the residual survival function
\[
S_R^{(k)}(t)=\mathbb{P}(R^{(k)}>t),
\]
where \(R^{(k)}\) is the remaining lifetime observed by the robot when it
encounters an obstacle in steady state. These two curves differ because
longer-lasting obstacles are more likely to be present when the robot arrives.
Therefore, the robot learns and plans using residual clearance times, not raw
clearance times from obstacle spawn. For a renewal process, the residual
survival function is
\[
S_R^{(k)}(t)
=
\frac{1}{\mathbb{E}[C^{(k)}]}
\int_t^\infty S_C^{(k)}(u)\,du .
\]

\subsection{Poisson Obstacle Spawning}

Obstacles are spawned independently of the robot according to a global Poisson
process. At each spawn event, the simulator samples an edge uniformly from the
set of blockable graph locations, samples an obstacle class, and then samples an
obstacle lifetime from the corresponding class-conditioned distribution. If the
selected edge is already occupied, the spawn is ignored, so each edge has at
most one active obstacle at a time.

The global spawn rate is chosen to achieve a desired steady-state blocked-edge
probability. Let \(m\) be the number of blockable undirected edges or spawn
locations, and let \(\lambda\) be the global obstacle spawn rate. Since edges are
selected uniformly, the per-edge spawn rate is
\[
\alpha = \frac{\lambda}{m}.
\]
Let \(\bar{C}\) be the mean obstacle duration averaged over the spawn
distribution,
\[
\bar{C} = \sum_{k\in\mathcal{K}} q_k\,\mathbb{E}[C^{(k)}],
\]
where \(q_k\) is the probability that a newly spawned obstacle has class \(k\).
Under the at-most-one-obstacle-per-edge model, each edge alternates between a
free period with mean \(1/\alpha\) and a blocked period with mean \(\bar{C}\).
Thus, the long-run blocked fraction is
\[
p_{\mathrm{block}}
=
\frac{\bar{C}}{1/\alpha+\bar{C}}
=
\frac{\alpha\bar{C}}{1+\alpha\bar{C}}.
\]
Solving for \(\alpha\) gives
\[
\alpha
=
\frac{p_{\mathrm{block}}}{\bar{C}(1-p_{\mathrm{block}})},
\]
and therefore
\[
\lambda
=
m\alpha
=
\frac{m p_{\mathrm{block}}}{\bar{C}(1-p_{\mathrm{block}})}.
\]
In the reported experiments, we set
\[
p_{\mathrm{block}}=0.05,
\]
so that approximately \(5\%\) of blockable edges are occupied in steady state.
The simulator is initialized with a warm-up period of \(1500\,\mathrm{s}\) before
the robot begins navigating, so that the obstacle field is close to its
steady-state distribution rather than starting from an empty environment.


\subsection{Class Spawn Weights}

The probability that an encountered obstacle has class \(k\) is not equal to the
probability that a newly spawned obstacle has class \(k\). Longer-lasting
obstacles remain in the environment for more time and are therefore encountered
more often. If \(q_k\) is the spawn probability and \(r_k\) is the desired
encounter probability, then
\[
r_k
=
\frac{q_k\mathbb{E}[C^{(k)}]}
     {\sum_j q_j\mathbb{E}[C^{(j)}]}.
\]
To obtain a desired encounter distribution \(r_k\), we choose spawn
probabilities proportional to
\[
q_k \propto \frac{r_k}{\mathbb{E}[C^{(k)}]},
\]
and then normalize. In simulation, the desired encounter mix is approximately
\[
r_{\mathrm{person}}=0.55,\qquad
r_{\mathrm{chair}}=0.30,\qquad
r_{\mathrm{bin}}=0.10,\qquad
r_{\text{tube}}=0.05.
\]
The unnormalized spawn weights are
\[
2626,\quad 119,\quad 28,\quad 8
\]
for person, chair, bin, and tube, respectively. After normalization, these
correspond to spawn probabilities of approximately
\[
94.4\%,\quad 4.3\%,\quad 1.0\%,\quad 0.3\%.
\]
Thus, people are spawned most frequently and tubes are spawned rarely, but
because tubes persist much longer, the encountered-obstacle distribution
matches the desired class mix.

\subsection{Learning and Planning Settings}

The learned policy begins without seeded censored or uncensored clearance
observations. Kaplan--Meier estimates are updated once per episode using all
observations collected during that episode. Thus, the survival model remains
fixed within an episode and is updated only after the episode terminates. This
matches the online learning setting in which the robot improves across repeated
navigation attempts.

For the wait-or-reroute decision, we evaluate candidate patience thresholds on a
grid. Both the patience-threshold grid and the downstream time grid use 300
points. The global maximum patience threshold is
\[
W_{\max}=2000\,\mathrm{s}.
\]
We also impose class-specific maximum patience thresholds:
\[
W_{\max}^{\mathrm{person}}=300\,\mathrm{s},
\qquad
W_{\max}^{\mathrm{chair}}=1000\,\mathrm{s},
\qquad
W_{\max}^{\mathrm{bin}}=1000\,\mathrm{s},
\qquad
W_{\max}^{\text{tube}}=1000\,\mathrm{s}.
\]
No explicit exploration is used in the reported experiments.

\subsection{Episodes, Timeouts, and Baselines}

Each simulated episode consists of one traversal from the start vertex to the
goal vertex. We use an episode timeout of 3600s.
If a policy fails to reach the goal before this timeout, the episode is treated
as unsuccessful and incurs the timeout duration in the time-to-goal statistics.
This convention is especially important for Greedy CTP. Since Greedy CTP
permanently removes blocked edges after observing them, it can eliminate useful
routes and sometimes make the remaining graph effectively disconnected or highly
inefficient. The large timeout therefore explains the poor mean time-to-goal
reported for Greedy CTP.

For the reported simulation results, we run 100 independent random seeds, with 1000 episodes per run.

\subsection{Scenario Manifests for Real-World Experiments}

The simulator also supports a sim-to-real mode used to generate real-world
experimental scenarios. In this mode, a simulation invocation writes a scenario
manifest for a future episode, including obstacle classes, spawn locations, and
timing information. These saved manifests allow the same obstacle scenario to be
recreated during physical robot experiments. After each real-world episode, the
field logs can be used to update the Kaplan--Meier survival estimates before
generating the next scenario.

\subsection{Generalization Across Graph Topologies and Blockage Rates}
\label{app:generalization}

To evaluate generalization beyond the original simulation setting, we consider three graph topologies: the original graph, a warehouse/corridor graph with expensive detours, and a grid with inexpensive alternative routes. For each topology, we vary the steady-state edge blockage probability over $p_{\mathrm{block}}\in\{0.02,0.05,0.10\}$. Results are averaged over the final 100 episodes of 10 random seeds. To compare performance across environments, we define normalized excess time as $D_{\mathrm{norm}}=(T-T_{\mathrm{free}})/T_{\mathrm{free}}$, where $T_{\mathrm{free}}$ is the obstacle-free shortest-path
time and $T$ is the actual time-to-goal.

Across all nine configurations, OSCAR remains close to oracle performance. On the original graph at $p_{\mathrm{block}}=0.10$, OSCAR achieves a mean time-to-goal of 268.8\,s, compared with 311.2\,s for Always Wait and 512.5\,s for Always Reroute.
Moreover, 12\% of OSCAR episodes finish within $3T_{\mathrm{free}}$, compared with only 3\% for Always Reroute. On the warehouse graph at the same blockage probability, Always Reroute incurs a normalized excess time of 60.6, compared with approximately 11.2 for OSCAR, the oracle, and Always Wait. On the grid, immediate rerouting remains competitive at low blockage probabilities because alternative
routes are inexpensive, but OSCAR approaches oracle performance as congestion increases. These results demonstrate that OSCAR adapts effectively across environments with different waiting and detouring costs.
\newpage
\section{Obstacle Classification Prompt}
\label{app:vlm_prompt}

In the real-world experiments, obstacle classes are obtained by querying a vision--language model (VLM) with a LiDAR reflectivity image and an aligned binary mask. The model is used for \emph{labeling only}; the navigation stack (not the VLM) decides whether to wait or reroute. The prompt is designed to be constrained to a fixed label set, focused on the masked obstacle region, and parsed from a deterministic output format.

The implementation settings are:
\begin{itemize}
    \item model: \texttt{gpt-5.3-chat-latest},
    \item API: OpenAI \texttt{responses.create} (Responses API),
    \item developer message: deterministic classification instructions (see below),
    \item \texttt{max\_output\_tokens}: \(200\),
    \item \texttt{store}: \texttt{true},
    \item mask mode: \texttt{binary} (default; white = obstacle, black = background),
    \item reflectivity topic: \texttt{/ouster/reflec\_image},
    \item duration estimation: disabled (\texttt{use\_vlm\_duration=false}); output is the obstacle label only.
\end{itemize}

The VLM is queried only when the active strategy requires classification (\textsc{RuleBased} or \textsc{Learned}); other strategies do not call the model. At runtime, if no mask is available, the implementation may omit the mask image and provide only the reflectivity image while keeping the same text instructions.

The exact messages used in our implementation are reproduced below.

\begin{verbatim}
Developer:
You are a deterministic robotics assistant for obstacle classification.
Follow instructions exactly and be consistent.

User:
You are an expert in indoor mobile robot navigation (Clearpath Jackal).
You are given a LiDAR intensity image (reflectivity/signal) and a
separate mask image highlighting the obstacle region. Focus ONLY on the
masked obstacle region(s). If there is both a person and another
obstacle like a chair or door, the other obstacle (chair/door) is the
dominant obstacle.

Second image is a binary mask aligned with the LiDAR image, where white
pixels mark the obstacle region and black is background. Focus ONLY on
the object in the white region.

Classify the dominant obstacle in the masked region by choosing EXACTLY
ONE label: {person, chair, sonotube, bin}.
A sonotube is a portable upright cardboard cylindrical column (a movable
fake pillar) on the floor.

Output EXACTLY: Obstacle: <label>.
\end{verbatim}

\begin{figure}[h]
    \centering
    \includegraphics[width=0.75
    \linewidth]{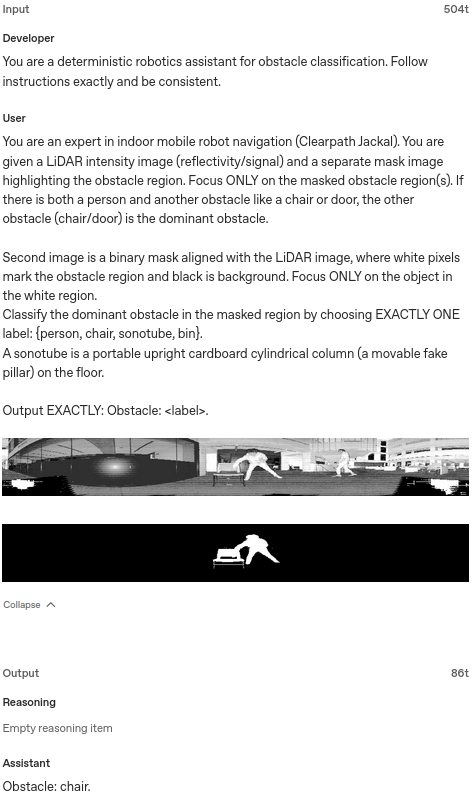}
    \caption{Example obstacle-classification log. The VLM receives a LiDAR
    reflectivity image and an aligned binary obstacle mask, and returns a
    deterministic class label in the required parseable format.}
    \label{fig:vlm_prompt_log}
\end{figure}

This prompt follows three design principles. First, it explicitly restricts attention to the masked region so that irrelevant scene content does not influence the classification. Second, it constrains the output to a small closed vocabulary matching the four obstacle types used in simulation and planning (person, chair, bin, and sonotube). Third, it enforces a fixed parseable output format (\texttt{Obstacle: <label>}), making the classifier easy to integrate into the navigation stack. An example query log is shown in Fig.~\ref{fig:vlm_prompt_log}, including the
reflectivity image, the aligned obstacle mask, and the parsed ChatGPT output used by the navigation stack.

\newpage
\section{Person/Chair-Only Real-World Results}
\label{app:unorchestrated_person_chair}

The real-world experiments in Sec.~\ref{realworldexp} use
orchestrated obstacle scenarios generated from the simulator, where obstacle
placement and removal times are specified in advance. We also evaluate the
learned policy in a more natural person--chair setting, where obstacles are not
placed or removed according to a timer. Instead, people and chairs appear and
clear according to the normal flow of activity in the environment. This provides
a less controlled but more natural test of whether the learned wait-or-reroute
behavior remains sensible outside the orchestrated scenario protocol.

\begin{figure}[h]
    \centering

    \begin{minipage}{0.32\linewidth}
        \centering
        \includegraphics[width=\linewidth]{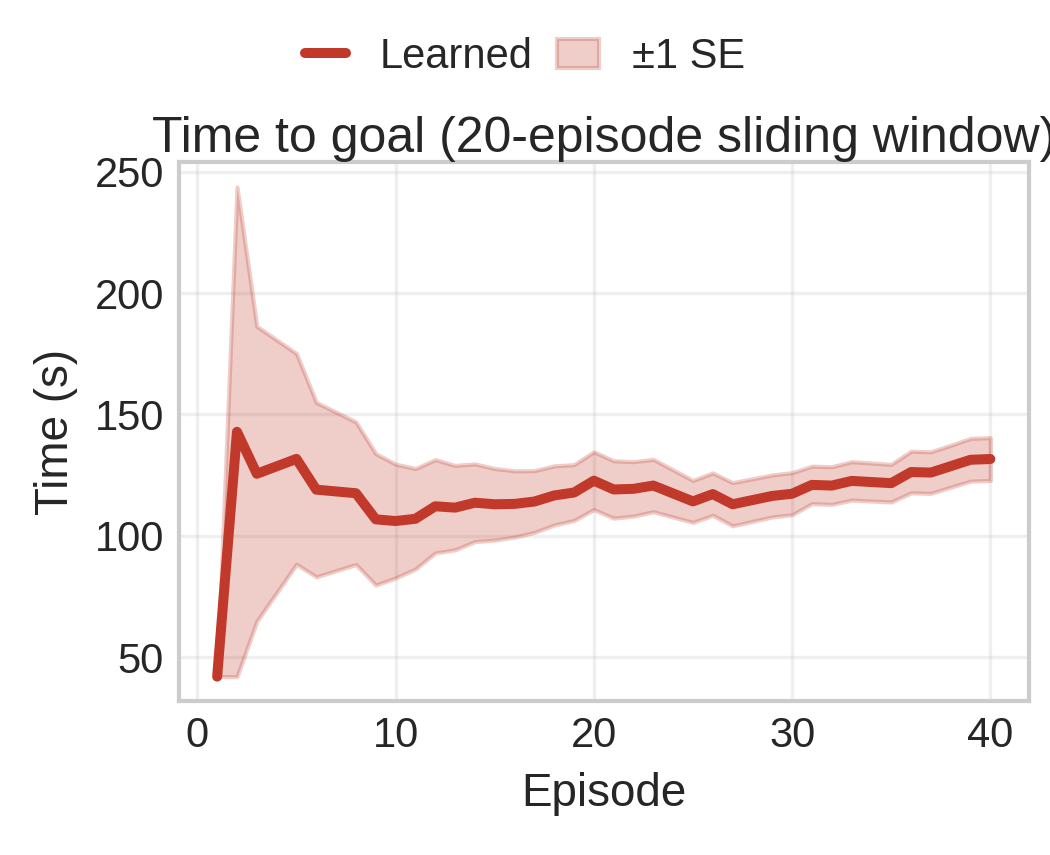}
        \vspace{-0.5em}
        \centerline{\small (a) Time-to-goal}
    \end{minipage}
    \hfill
    \begin{minipage}{0.32\linewidth}
        \centering
        \includegraphics[width=\linewidth]{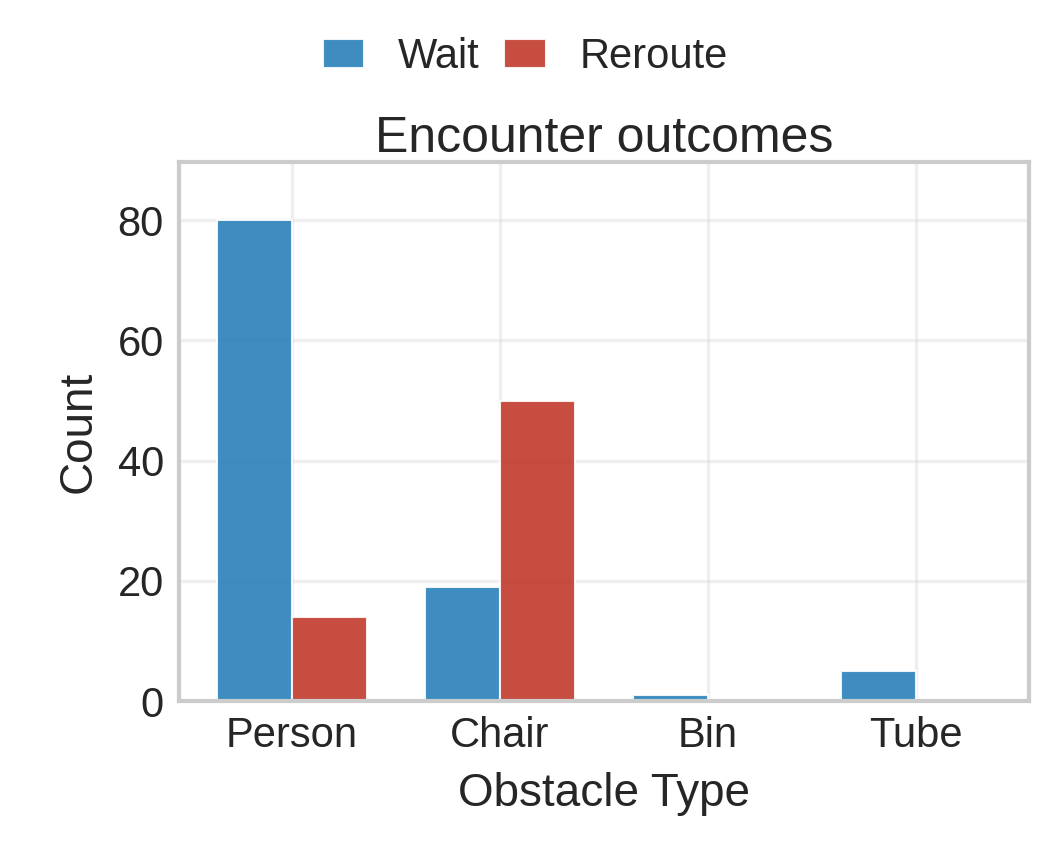}
        \vspace{-0.5em}
        \centerline{\small (b) Encounter outcomes}
    \end{minipage}
    \hfill
    \begin{minipage}{0.32\linewidth}
        \centering
        \includegraphics[width=\linewidth]{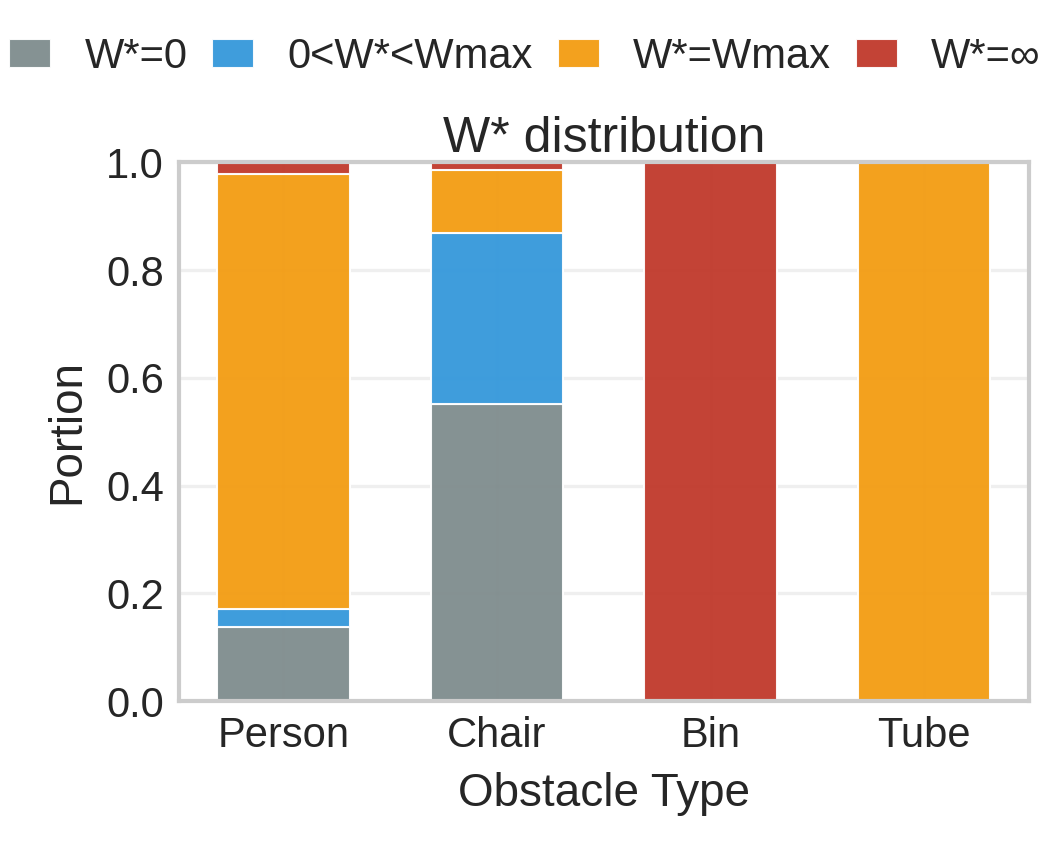}
        \vspace{-0.5em}
        \centerline{\small (c) Patience thresholds}
    \end{minipage}

    \caption{Unorchestrated person--chair real-world experiment. Unlike the
    orchestrated real-world scenarios, obstacles are not introduced or removed
    according to a prescribed timer. The learned policy usually waits for
    people, more often reroutes around chairs, and selects patience thresholds
    consistent with these behaviors. Bin and sonotube entries correspond to
    ChatGPT obstacle-labeling misclassifications in an environment containing
    only people and chairs.}
    \label{fig:field_person_chair_results}
\end{figure}

Figure~\ref{fig:field_person_chair_results} summarizes the results. The
time-to-goal initially varies substantially, but stabilizes after several
episodes as the policy accumulates experience. Note that the increase at the end of Figure~\ref{fig:field_person_chair_results}(a) can be credited to an increase in the number of obstacles in our scene as episode count increases. The encounter outcomes show that
the robot usually waits for people, which are typically transient, while chairs
lead to a more reroute-heavy response. This behavior is also reflected in the
selected patience thresholds: person encounters are usually assigned
\(W^\star=W_{\max}\), while chair encounters more often receive either immediate
reroutes or finite patience thresholds. Although the environment only contains
people and chairs, a small number of detections are labeled as bin or sonotube;
these are misclassifications made by ChatGPT during obstacle labeling, not
additional obstacle classes present in the experiment.

\end{document}